\pdfoutput=1

\documentclass[11pt]{article}

\usepackage[]{acl}

\usepackage{times}
\usepackage{graphicx}
\usepackage{latexsym}
\usepackage{multirow}
\usepackage{multicol}
\usepackage{subcaption}
\usepackage{xcolor}
\usepackage{mdframed}

\usepackage[T1]{fontenc}

\usepackage[utf8]{inputenc}
\usepackage{url}
\usepackage{microtype}

\title{LexSumm and LexT5: Benchmarking and Modeling \\ Legal Summarization Tasks in English}

\author{ Santosh T.Y.S.S, \bf{Cornelius Weiss, Matthias Grabmair} \\ School of Computation, Information, and Technology; \\
Technical University of Munich, Germany \\ \ }

\begin{document}
\maketitle
\begin{abstract}
In the evolving NLP landscape, benchmarks serve as yardsticks for gauging progress. However, existing Legal NLP benchmarks only focus on predictive tasks, overlooking generative tasks. This work curates LexSumm, a benchmark designed for evaluating legal summarization tasks in English. It comprises eight English legal summarization datasets, from diverse jurisdictions, such as the US, UK, EU and India. Additionally, we release LexT5, legal oriented sequence-to-sequence model, addressing the limitation of the existing BERT-style encoder-only models in the legal domain. We assess its capabilities through zero-shot probing on LegalLAMA and fine-tuning on LexSumm. Our analysis reveals abstraction and faithfulness errors even in summaries generated by zero-shot LLMs, indicating opportunities for further improvements. LexSumm benchmark and LexT5 model are available at \url{https://github.com/TUMLegalTech/LexSumm-LexT5}.
\end{abstract}

\section{Introduction}
Language serves as the bedrock of the legal domain, facilitating precise communication in this complex field. Legal systems globally engage in the production, consumption and interpretation of massive volumes of text. Legal professionals, comprising lawyers, judges and regulators, continually author a diverse array of complex legal documents, such as briefs, memos, statutes, regulations, contracts, patents and judicial decisions \cite{coupette2021measuring}. In their routines, these professionals not only craft these documents but also immerse themselves in extensive volumes of text, refining their comprehension of the law for effective human behavior management. Beyond the realms of consumption and production, the practice of law and the art of lawyering hinge on the analysis and interpretation of textual content \cite{chalkidis2022lexglue}, often perceived by laypersons as legalese or legal gobbledygook \cite{katz2023natural}.

Recent advancements in NLP stand poised to revolutionize legal tasks and significantly benefit stakeholders within the legal domain \cite{zhong2020does}.  By automating labor-intensive processes, such as document analysis \cite{wang2023maud,koreeda2021contractnli,lippi2019claudette,graham2023natural,sancheti2023read}, information extraction \cite{luz2018lener,chen2020joint,hendrycks2021cuad,chalkidis2017extracting}, question answering \cite{ravichander2019question,kien2020answering,zhong2020iteratively,zhong2020jec,chen2023equals,louis2023interpretable,zheng2021does}, text classification \cite{chalkidis2019large,chalkidis2021multieurlex,tuggener2020ledgar,santosh2024chronoslex}, information retrieval \cite{louis2022statutory,ma2021lecard,shao2020bert,santosh2024ecthr,santosh2024query} and summarization \cite{shukla2022legal,bhattacharya2019comparative,bhattacharya2021incorporating,schraagen2022abstractive,elaraby2022arglegalsumm,elaraby2023towards,zhong2019automatic,xu2021toward,xu2023argumentative,santosh2024beyond,tyss2024lexabsumm}, NLP with its ability to understand and interpret complex legal language can enhance efficiency and accelerate decision-making. NLP can act as a force multiplier by not only streamlining tasks but also amplifiying the capabilities of legal professionals, leading to increased productivity of legal stakeholders \cite{katz2023natural}.

To enable a systematic comparison of approaches, legal evaluation benchmarks like LexGLUE \cite{chalkidis2022lexglue} and LEXTREME \cite{niklaus2023lextreme} have been proposed, focusing on predictive tasks. However, there is an absence of a dedicated benchmark designed for assessing legal generation capabilities. Moreover, resources on Legal Natural Language Generation (NLG) are sporadic and scattered. In response to this, we introduce LexSumm, a new benchmark curated for training and evaluating legal English summarization models. It includes eight English legal summarization datasets from various jurisdictions, such as the US, UK, EU, and India, for training task-specific models—distinguishing it from LegalBench \cite{guha2023legalbench} and LawBench \cite{fei2023lawbench}, oriented towards zero/few-shot LLM evaluation.

LexSumm represents the distinctive characteristic of legal documents, marked by their long length, posing a challenge for pre-trained models like BART \cite{lewis2020bart} and T5 \cite{raffel2020exploring}. In our benchmarking efforts, we evaluate LexSumm using long-context models such as LED \cite{beltagy2020longformer}, LongT5 \cite{guo2022longt5}, and PRIMERA \cite{xiao2022primera}. We also explore contemporary approaches of adopting short-range pre-trained models like T5 \cite{raffel2020exploring} with fusion-in-decoder techniques as in SLED \cite{ivgi2023efficient} and integration of retrieval techniques, as demonstrated in Unlimiformer \cite{bertsch2023unlimiformer}, to adopt them for longer documents. Additionally, we compare recent long-context zero-shot LLMs like GPT-3.5 and Claude on LexSumm.

Pre-trained language models such as BERT \cite{kenton2019bert}, RoBERTa \cite{liu2019roberta} have significantly transformed the NLP landscape, showcasing remarkable efficacy in general-domain text. However, their performance diminishes when applied to domain-specific tasks, leading to concept of continued pre-training with domain-specific unlabeled data \cite{gururangan2020don}. This resulted in the development of legal-specific pre-trained models like LegalBERT \cite{chalkidis2020legal, zheng2021does, henderson2022pile, chalkidis-etal-2023-lexfiles}. To the best of our knowledge, there has been a lack of sequence-to-sequence model tailored for the legal domain. To address this gap, we introduce LexT5, an English legal-oriented sequence-to-sequence model pre-trained on the LeXFiles corpus \cite{chalkidis-etal-2023-lexfiles}, 
from six English-speaking legal systems (EU, European Council, Canada, US, UK, India). To evaluate the legal knowledge acquired by LexT5, we compare its to T5 on LegalLAMA \cite{chalkidis-etal-2023-lexfiles}, a zero-shot legal probing suite. We also assess LexT5's performance on LexSumm by incorporating into SLED and Unlimiformer frameworks to accommodate longer inputs. 

Our quantitative and qualitative analysis reveal that LexSumm presents a substantial challenge for existing models including zero-shot LLMs such as GPT-3.5, leaving much room for the research community to improve upon. To streamline future model evaluations, we will release our benchmark and our pre-trained LexT5 model on the Hugging Face Hub, contributing to the advancement of legal NLP research.

\section{Related Work}
\noindent \textbf{NLG benchmarks}
\citealt{liu2021glge} introduced GLGE, a benchmark focusing on English NLG with eight datasets across four tasks. For Chinese, there are CUGE \cite{yao2021cuge} and LOT \cite{guan2022lot}, with both language understanding and generation tasks. BanglaNLG \cite{bhattacharjee2023banglanlg} serves as a generation benchmark for Bangla with seven datasets across six tasks. Dolphin \cite{elmadany2023dolphin} offers a comprehensive benchmark for Arabic NLG, covering 13 different tasks. GEMv1 \cite{gehrmann2021gem} is a multilingual NLG benchmark spanning 18 languages and 13 datasets. It has been extended with GEMv2 \cite{gehrmann2022gemv2}, encompassing 51 languages. IndoNLG \cite{cahyawijaya2021indonlg} focuses on 3 Indonesian languages, while IndicNLG \cite{kumar2022indicnlg} covers 11 Indic languages. MTG \cite{chen2022mtg} spans 5 languages. 

Turning to specific domains, MedEval \cite{he2023medeval} and M3 \cite{otmakhova2022m3} are benchmarks tailored for the medical domain, with classification and generation tasks. In line with these efforts, this work introduces LexSumm, a legal domain-specific summarization benchmark with eight datasets.
\vspace{1mm}

\noindent \textbf{Benchmarks for Legal Domain}
LexGLUE \cite{chalkidis2022lexglue} stands out as the pioneering benchmark in the legal domain, evaluating NLP models on tasks related to legal language understanding. It encompasses seven classification tasks derived from six English legal NLP datasets, spanning jurisdictions such as the US, EU, and the Council of Europe. LEXTREME \cite{niklaus2023lextreme} is a multilingual benchmark for the legal domain, comprising 11 relevant NLU datasets covering 24 languages 
from two language families (Indo- European and Uralic). 
FairLex \cite{chalkidis2022fairlex}, another legal benchmark focuses on assessing fairness across five attributes—gender, age, region, language, and legal area—across legal NLP tasks. FairLex covers four jurisdictions (European Council, USA, Switzerland, and China), supports five languages (English, German, French, Italian, and Chinese). LBOX \cite{hwang2022multi} benchmarks Korean legal tasks, consisting of two classification tasks, two legal judgment prediction tasks, and one summarization task. LegalBench \cite{guha2023legalbench} is construced to assess legal reasoning consisting of 162 tasks covering six different types of legal reasoning, designed for benchmarking zero/few-shot LLM paradigm for English language primarily based on American laws.  Similarly, LawBench \cite{fei2023lawbench} is LLM oriented benchmark designed for assessing chinese civil-law system, containing 20 diverse tasks covering 5 task types: single-label, multi-label classification, regression, extraction and generation. 

In this work, we curate LexSumm benchmark, focusing on eight legal summarization datasets in English, facilitating fine-tuning of task-specific models, an important setting for numerous applications. LexSumm Benchmark, with generative tasks, complements the LexGLUE benchmark \cite{chalkidis2022lexglue} for legal text understanding in English. 
\vspace{1mm}

\noindent \textbf{Legal Pre-trained models}
\citet{gururangan2020don} demonstrated continuing pre-trained models on domain-specific text improves performance on domain tasks. Subsequently, there have been efforts to continue pre-training on diverse English legal text like legislation, court cases, and contracts, spanning US, EU, and UK jurisdictions resulting in the creation of LegalBERT \cite{chalkidis2020legal}. In a similar vein, CaseLawBERT \cite{zheng2021does} is another law-specific BERT model trained using the Harvard Law case corpus from US federal and state courts. \citet{henderson2022pile} compiled an extensive corpus known as Pile of Law, incorporating documents from the US, Canada, and EU and trained BERT-large on this corpus, giving rise to the PoLBERT. \citealt{paul2023pre} extended pre-training on the Indian legal corpora, culminating in InLegalBERT. Recently, \citealt{chalkidis-etal-2023-lexfiles} introduced LexLMs, pre-trained on LeXFiles, a diverse multinational English legal corpus from six primarily English-speaking legal systems. 

While the aforementioned models focus on English legal corpora, parallel endeavors have emerged to develop legal pre-trained models other languages. French legal model, JuriBERT \cite{douka2021juribert} is trained using corpora from the Court of Cassation, France's highest court. Similar initiatives include JurBERT for Romanian \cite{masala2021jurbert}, LamBERTa, ItalianLegalBERT for Italian \cite{tagarelli2022lamberta, licari2022italian}, RoBERTalex for Spanish \cite{gutierrez2021spanish}, Lawformer for Chinese \cite{xiao2021lawformer}, AraLegalBERT for Arabic \cite{al2022aralegal},  LCUBE for Korean \cite{hwang2022multi}, and LegalBERT-pt, BERTBR for Portuguese \cite{ciurlino2021bertbr}. Recently, \citet{niklaus2023multilegalpile} introduced LegalXLM, a multilingual model pre-trained on the MultiLegalPile, a diverse legal corpus comprising 24 languages from 17 jurisdictions.

It is noteworthy that the aforementioned legal domain-specific pre-trained language models predominantly adhere to the BERT-style encoder-only architecture and currently, there is a lack of sequence-to-sequence models specifically adapted for legal text. Addressing this gap, we present LexT5, legal-oriented sequence-to-sequence model pre-trained on the LexFiles corpus for English.

\section{LexSumm Benchmark}
LexSumm Benchmark is a collection of eight legal NLG datasets 
in English language spanning across US, EU, UK and India jurisdictions. In this section, we describe these datasets and their characteristics.  

\vspace{1mm}
\noindent \textbf{BillSum} \cite{kornilova2019billsum} is a summarization dataset of US Congressional bills, sourced from the Govinfo service 
by the US Government Publishing Office 
along with human-written summary from the Congressional Research Service
. It consists 
of 22218 document-summary pairs split into training (16664), validation (2222) and test (3322) sets. 

\vspace{1mm}
\noindent \textbf{EurLexSum} \cite{aumiller-etal-2022-eur} EUR-Lex platform provides access to various legal documents 
published by various European Union organs. This dataset focuses on the enforced EU legislation 
along with their summaries
, available  across all 24 european languages. 
We restrict to English version of the dataset spanning 1504 document-summary pairs, split into 1128/151/225 for training, validation and testing respectively. 

\vspace{1mm}
\noindent \textbf{GovReport} \cite{huang2021efficient}  contains 19,465 national policy reports published by U.S. Government Accountability Office 
An expert-written summary is provided along with each report and it is split into 14598, 2919, 1946 for training, validation and test sets respectively.

\vspace{1mm}
\noindent \textbf{MultiLexSum-Tiny/Short/Long} \cite{shen2022multi} consists of 9280 expert-written summaries for 4500 documents from U.S. federal civil rights lawsuits. It has summaries at three different granularities for the same source: (a) Long (L) summaries contain multiple paragraphs, covering the case background, parties involved, major case events and proceedings. 
(b) Short (S) summaries have only one paragraph with a shorter description of the background, parties involved and the outcome of the case. (c) Tiny (T) summaries have one sentence intended to appear on Twitter. 
Input spans across multiple sources such as first complaint, last amended complaint, settlement agreements, 
opinions, orders etc. 
Three different summarization tasks at each granularity are proposed emulating real-world tasks at the Civil Rights Litigation Clearinghouse.
Long, Short and Tiny versions have a total of 4539, 3138 and 1603 document-summary pairs respectively which  are split into (3404/454/681), (2340/312/486) and (1207/145/251) for train, validation and test. 

\vspace{1mm}
\noindent \textbf{InAbs} \cite{shukla2022legal} consists of Indian Supreme Court judgements collected from the website of Legal Information Institute of India 
. It provides summaries (also called `headnotes') for some of the cases resulting in total of 7150 case document-summary pairs, which are split into training (5346), validation (713) and test (1069) sets.  

\vspace{1mm}
\noindent \textbf{UKAbs} \cite{shukla2022legal} dataset is collected from the UK Supreme court website
which provides all judgements that were ruled since 2009. For most of the cases, along with the judgements, it also provides the official press summary of the cases. It consists of 793 document-summary pairs which are split into 595, 79, 119 for training, validation and test respectively.

\begin{table*}[]
\centering
\scalebox{0.75}{
\begin{tabular}{|l|c|c|c|c|c|c|c|c|}
\hline
 & \textbf{BillSum} & \textbf{EurLexSum} & \textbf{GovReport} & \textbf{MLS-Long} & \textbf{MLS-Short} & \textbf{MLS-Tiny} & \textbf{INAbs} & \textbf{UKAbs} \\ \hline
Input Len        & 1665.14          & 16390.28           & 8765.03            & 75255.36          & 99460.62           & 118347.65         & 4839.76        & 15911.07       \\ 
Summary Len      & 204.09           & 960.46             & 556.31             & 639.18            & 128.63             & 25.19             & 941.58         & 1240.75        \\ 
Comp. Ratio      & 13.21            & 17.29              & 17.83              & 98.82             & 874.18             & 5681.723          & 5.97           & 12.65          \\ 
Coverage@1       & 0.89             & 0.87               & 0.94               & 0.93              & 0.95               & 0.92              & 0.94           & 0.96           \\ 
Coverage@2       & 0.58             & 0.53               & 0.67               & 0.61              & 0.65               & 0.51              & 0.76           & 0.67           \\ 
Density@1        & 3.89             & 6.11               & 9.27               & 4.07              & 3.33               & 2.26              & 13.99          & 9.91           \\ 
Density@2        & 2.61             & 4.89               & 8.09               & 2.93              & 2.21               & 1.18              & 12.67          & 8.66           \\ \hline
\end{tabular}}
\caption{Characteristics of eight datasets in LexSumm. MLS, Len denote MultiLexSumm and length respectively.}
\label{data-char}
\end{table*}

\subsection{Dataset Characteristics}
We report the following characteristics on the eight datasets of LexSumm in Table \ref{data-char}.

\noindent (a) Average number of words in the input text and the summary. 
We also plot the token length distribution for the input and summary in Fig. \ref{fig:data-char1} and \ref{fig:data-char2}. (b) Compression Ratio \cite{grusky2018newsroom} indicates the token ratio between the input to the summary. (c) Coverage@n \cite{grusky2018newsroom} quantifies the extent to which a summary is derivative of a input text. It indicates the ratio of n-grams in the summary that are part of an extractive fragment within the input. (d) Density@n \cite{grusky2018newsroom} quantifies how well the n-gram sequence of a summary can be described as a series of extractions. It is defined as the average length of the extractive fragment to which each n-gram in the summary belongs. For instance, a summary might contain many individual n-grams from the input indicating a high coverage. However, if dispersed across the input (less density), these n-grams of the summary could still be used in abstractive sense and not merely extractive from the article. 
(e) Fusion score \cite{shaham2022scrolls} measures how the summary sentences are synthesized from multiple sentences or compressed from a single sentence in the input. We plot the distribution of fusion score in Fig. \ref{fig:data-char1} and \ref{fig:data-char2}, by computing fusion spread score for each instance as the standard deviation between the locations of output bigrams in the input (if exists). 

We observe that LexSumm encompasses datasets with a diverse range of input-output lengths, leading to varying compression ratios. MultiLexSumm, with its three different granularities, exhibits higher compression ratios, indicating the need to precisely capture the critical aspects of the input text, highlighting its challenging nature. Although the coverage@1 scores for all datasets exceed 0.8, indicating fewer novel terms introduced into the summary (less paraphrasing involved), hinting at the extractive nature. However, the bi-gram coverage is lower, indicating that these extractive tokens are dispersed across the input, resulting in less density  and larger fusion spread in Fig. \ref{fig:data-char1} and \ref{fig:data-char2}. INAbs emerges as the most extractive dataset with a smaller compression ratio and higher coverage and density values, followed by UKAbs and GovReport. Conversely, MultiLexSumm, with its higher compression ratio, lower coverage and density values, emerges as the most abstractive dataset.

\section{LexT5}
We build LexT5, a legal-specific seq2seq pre-trained model. 
T5 is an encoder-decoder model initially pre-trained in an unsupervised manner on the 
C4 corpus \cite{raffel2020exploring}, using span denoising objective which
involves replacing 15\% of the tokens with sentinel tokens 
along with consecutive tokens marked for removal being replaced by a single sentinel token. The resulting corrupted text serves as input to the model to predict the masked-out span. Then the model is further fine-tuned using supervised training on various downstream tasks, including those from the GLUE and SuperGLUE \cite{wang2018glue, wang2019superglue} benchmarks, casting them into text-to-text format for training. 

We initialize the model with T5-base 
checkpoint of \citet{raffel2020exploring} and continue pre-training using the span denoising objective 
on the train split of LeXFiles  \cite{chalkidis-etal-2023-lexfiles}. LeXFiles is a diverse legal corpus across 6 primarily English-speaking legal systems (EU, European Court of Human Rights, Canada, US, UK, India) covering various legal documents such as legislation, case law and contracts. It comprises approx. 6 million documents totalling up to approx. 19 billion tokens. We employ a sentence sampling rate 
from each sub-corpora proportional to number of tokens with exponential smoothing factor of 0.5 \cite{liu2020multilingual}. 
Implementation details in App \ref{pre-impl}.

\begin{table*}[]
\centering
\scalebox{0.75}{
\begin{tabular}{|l|c|c|c|c|cc|cc|}
\hline
& \multirow{2}{*}{\#Inp} & \multirow{2}{*}{\begin{tabular}[c]{@{}c@{}}\#Tok/ \\ Inp\end{tabular}} & \multirow{2}{*}{\#Tgt} & \multirow{2}{*}{\begin{tabular}[c]{@{}c@{}}\#Tok/\\ Tgt\end{tabular}} & \multicolumn{2}{c|}{T5}          & \multicolumn{2}{c|}{LexT5}       \\ \cline{1-1} \cline{6-9} 
Tasks                           &                        &                                                                            &                        &                                                                            & \multicolumn{1}{c|}{NLL $\downarrow$ }  & MRR $\uparrow$  & \multicolumn{1}{c|}{NLL $\downarrow$ }  & MRR $\uparrow$  \\ \hline
Articles (ECHR)                 & 5063                   & 147.67                                                                     & 13                     & 1                                                                          & \multicolumn{1}{c|}{1.77} & 0.45 & \multicolumn{1}{c|}{0.31} & 0.93 \\ 
Contractual Sec. Titles (US) & 1527                   & 224.58                                                                     & 20                     & 2.5                                                                        & \multicolumn{1}{c|}{1.97} &  0.64    & \multicolumn{1}{c|}{1.44} & 0.71 \\ 
Contract Types (US)             & 1062                   & 149.34                                                                     & 15                     & 1.4                                                                        & \multicolumn{1}{c|}{4.63} & 0.38 & \multicolumn{1}{c|}{2.87} & 0.68 \\ 
Crime Charges (US)              & 4518                   & 276.99                                                                     & 116                    & 3.28                                                                       & \multicolumn{1}{c|}{1.9}  & 0.49 & \multicolumn{1}{c|}{1.67} & 0.56 \\ 
Legal Terminology (US)          & 5806                   & 286.04                                                                     & 145                    & 3.13                                                                       & \multicolumn{1}{c|}{2.58} & 0.53 & \multicolumn{1}{c|}{1.74} & 0.74 \\ 
Legal Terminology (EU)          & 2127                   & 160.92                                                                     & 53                     & 3.49                                                                       & \multicolumn{1}{c|}{2.38} & 0.55 & \multicolumn{1}{c|}{0.91} & 0.83 \\ 
Legal Terminology (ECHR)        & 6273                   & 166.49                                                                     & 143                    & 3.36                                                                       & \multicolumn{1}{c|}{2.24} & 0.55 & \multicolumn{1}{c|}{0.78} & 0.88 \\ 
Criminal Code Sec. (Canada) & 321                    & 148.56                                                                     & 195                    & 3.42                                                                       & \multicolumn{1}{c|}{2.2}  & 0.33 & \multicolumn{1}{c|}{0.91} & 0.7  \\ \hline
\end{tabular}}
\caption{Data Characterstics  of LegalLAMA probing suite and NLL, MRR values for T5 and LexT5 models. \#Inp, \#Tok/Inp, \#Tgt, \#Tok/Tgt indicate number of test instances, average number of tokens per input, the number of target spans and the average number of tokens per target respectively.}
\label{probe-results}
\end{table*} 

\subsection{Probing Legal Knowledge}
To assess legal knowledge acquired by the model during pre-training phase, we use LegalLAMA \cite{chalkidis-etal-2023-lexfiles}, a legal concept probing benchmark suite 
similar to LAnguage Models Analysis (LAMA) probing suite \cite{petroni2019language}. The zero-shot probing task is defined as follows: Given a sentence with a masked span [mask], the model must predict the gold masked span. Unlike encoder-only models like BERT, which require multiple masks to predict multi-token targets, T5's pre-training strategy replaces consecutive masked tokens with a single mask token resulting in a more robust evaluation for the probing task. Note that LegalLAMA instances are derived from the test subset of LexFiles to prevent contamination from pre-training corpus.

LegalLAMA consists of 8 tasks: (i) Articles (ECHR): The model predicts the masked article number in paragraphs from ECtHR decisions. (ii) Contractual Section Titles (US): Predicting the masked section titles in US contracts. (iii) Contract Types (US): Predicting the masked contract type in introductory paragraphs of US contracts. (iv) Crime Charges (US): Predicting masked criminal charges in paragraphs from US court judgments. (v) Legal Terminology (US): Predicting masked legal terms based on vocabularies from the Legal Information Institute 
in paragraphs from US court judgments. (vi) Legal Terminology (EU): Predicting masked legal terms based on subject matters from the CURIA database in paragraphs from CJEU judgments. (vii) Legal Terminology (ECHR): Predicting masked legal terms or issues based on keywords from the HUDOC database in paragraphs from ECHR case documents. (viii) Criminal Code Sections (Canada): Predicting masked sections of the Criminal Code of Canada in paragraphs from Criminal Court of Canada decisions. 

Statistics about the test instances count, average input token count, target spans count and average tokens per target span for the eight tasks are presented in Table \ref{probe-results}. We calculate token-normalized negative log-likelihood (NLL) loss across the golden target span for each instance and report average across all instances. Lower NLL signifies a better aquisition of legal knowledge by the model. We also compute Mean Reciprocal Rank (MRR) \cite{voorhees1999trec} for each instance based on the ranking list over the set of candidate target spans and report the average across all instances. The ranking list is based on the increasing order of token-normalized NLL values. Higher MRR indicates a superior acquisition of legal knowledge, with an ideal value of 1.0.

We present the NLL and MRR values for both the T5 and LexT5 models in Table \ref{probe-results}. Across all tasks, we observe that LexT5 achieves lower NLL and higher MRR values compared to T5, indicating acquisition of legal knowledge through pre-training on the LeXFiles corpus. Notably, Crime Charges (US) and Contractual Section Titles (US) exhibit the smallest increase, with a marginal 0.07 MRR points, despite US being the dominant in LexFiles ($\approx$ 70\%). Surprisingly, we do not find a correlation between the target spans count and the average token count in target span with performance improvements, contradicting findings of \cite{chalkidis-etal-2023-lexfiles}, which observed an increase in performance negatively correlated with the average tokens count of target spans. We attribute this discrepancy to the probing design bias in encoder-only models, where the number of masks already encode a signal for the token count of the target span. In contrast, our setup ensures a more reliable approach by not leaking the number of tokens in the target span, as we only have one mask for the whole span.

\section{Benchmarking Experiments}
We benchmark 8 LexSumm tasks using the following seq2seq models,  designed to handle longer inputs. Implementation details are in App. \ref{impl-details}.
\vspace{1mm}

\noindent \textbf{LED} \cite{beltagy2020longformer} is based on Longformer, an efficient transformer model with linear complexity relative to input length. It features encoder and decoder components, employing efficient local+global attention in the encoder and full quadratic attention in the decoder. LED is initialized from pre-trained BART \cite{lewis2020bart}, with the position embedding matrix initialized by duplicating BART’s 1K position embeddings 16 times to handle 16k input tokens. 

\vspace{1mm}
\noindent \textbf{PRIMERA} \cite{xiao2022primera} is initialized with the LED model and pre-trained with a novel summarization-specific masking objective based on the entity pyramid evaluation method, inspired by the Gap Sentence Generation objective of Pegasus \cite{zhang2020pegasus}. It can handle 4096 tokens.

\begin{table*}[]
\centering
\scalebox{0.733}{
\begin{tabular}{lcccc}
 \hline
 \multicolumn{1}{c}{\textbf{}} & \textbf{R-1 / 2 / L / BS} & \textbf{R-1 / 2 / L / BS} & \multicolumn{1}{c}{\textbf{R-1 / 2 / L / BS}} & \textbf{R-1 / 2 / L / BS} \\ \hline
\multicolumn{1}{c}{\textbf{}} & \textbf{BillSum}          & \textbf{EurLexSumm}       & \textbf{GovReport}  & \textbf{MLS-Long}         \\ \hline
LED                           & \textbf{38.7} / 22.1 / 36.0 / 64.1 & 36.8 / 18.8 / 33.7 / 67.3 & 38.6 / 19.3 / 35.9 / 66.4 & 40.1 / \underline{20.4} / \underline{37.0} / 68.3 \\
PRIMERA                       & 37.0 / 21.7 / 35.5 / 63.6 & 32.7 / 16.8 / 30.8 / 64.8 &  37.8 / 19.0 / 35.1 / 65.5 & 38.2 / 19.0 / 35.4 / 67.6 \\
LongT5                        & 38.6 / 22.9 / 36.1 / 65.6 & 34.7 / 17.6 / 30.8 / 66.6 & 38.3 / 19.8 / 35.5 / 66.4 & 39.1 / 20.2 / 36.2 / 67.7 \\
SLED-T5                       & 36.8 / 22.9 / 35.2 / 64.8 & 36.5 / 18.7 / 33.2 / 67.0 & 38.4 / 19.7 / 35.4 / 66.1 & 38.6 / 19.5 / 35.5 / 66.2 \\
Unlim.- T5                    & 36.9 / 23.2 / 35.4 / 65.1 & 35.5 / 18.6 / 33.5 / 67.1 & 38.2 / 19.5 / 35.9 / 65.7 & 38.7 / 20.0 / 36.2 / 66.9 \\
SLED-LexT5                    & 38.2 / \underline{24.5} / \underline{36.1} / \underline{66.0} & \underline{37.4} / \textbf{19.3} / \textbf{34.3} / \textbf{67.6} & \underline{39.4} / \underline{20.7} / \underline{36.4} / \textbf{66.8} & \underline{40.4} / 19.8 / 36.5 / \underline{68.4} \\
Unlim.-LexT5                  & \underline{38.4} / \textbf{24.7} / \textbf{36.4} / \textbf{66.1} & \textbf{37.9} / \underline{19.1} / \underline{34.1} / \underline{67.5} & \textbf{40.2} / \textbf{21.2} / \textbf{36.9} / \underline{66.6} & \textbf{41.6} / \textbf{20.8} / \textbf{37.6} / \textbf{68.8} \\
\hline                                    
\multicolumn{1}{c}{\textbf{}} & \textbf{MLS-Short}        & \textbf{MLS-Tiny}         & \textbf{INAbs}            & \textbf{UKAbs}            \\ \hline
LED                           & 37.5 / 18.4 / 34.4 / 65.1 & 24.9 / 11.1 / 22.6 / 56.8 & \textbf{42.8} / 23.8 / 39.2 / 67.9 & \textbf{38.8} / 18.2 / 35.4 / 67.5 \\
PRIMERA                       & 36.4 / 18.2 / 33.5 / 64.0 & 24.5 / 10.8 / 22.5 / \underline{56.9} & 39.2 / 21.0 / 36.1 / 66.1 & 36.4 / 16.6 / 33.1 / 65.1 \\
LongT5                        & 37.7 / 18.0 / 34.6 / \textbf{65.6} & 24.4 / 10.3 / 22.0 / 56.5 & 40.6 / 21.4 / 36.8 / 66.6 & 36.1 / 17.3 / 33.4 / 66.1 \\
SLED-T5                       & 36.8 / 17.8 / 34.2 / 64.7 & 24.4 / 11.0 / 22.2 / 56.6 & 39.5 / 22.3 / 36.7 / 67.1 & 36.5 / 18.3 / 34.3 / 66.7 \\
Unlim.- T5                    & 36.3 / 17.6 / 34.1 / 64.4 & 25.2 / 11.1 / 23.7 / 56.7 & 40.0 / 22.7 / 37.1 / 67.2 & 37.5 / 18.2 / 34.2 / 66.9 \\
SLED-LexT5                    & \underline{38.4} / \underline{18.7} / \textbf{35.6} / \textbf{65.6} & \underline{26.3} / \underline{12.2} / \underline{23.7} / 56.8 & 41.1 / \underline{24.3} / \underline{39.5} / \underline{68.2} & \textbf{38.8} / \underline{18.8} / \underline{35.5} / \textbf{68.0} \\
Unlim.-LexT5                  & \textbf{38.8} / \textbf{19.1} / \textbf{35.6} / \underline{65.3} & \textbf{27.5} / \textbf{12.4} / \textbf{24.7} / \textbf{57.3} & \underline{42.2} / \textbf{24.5} / \textbf{39.7} / \textbf{68.4} & \underline{38.2} / \textbf{18.9} / \textbf{35.9} / \underline{67.9} \\
\hline
\end{tabular}}
\caption{Evaluation results of various  models across eight datasets of LexSumm. Best and second best value under each metrics is bolded and underlined respectively.}
\label{results}
\end{table*}

\vspace{1mm}
\noindent \textbf{LongT5} \cite{guo2022longt5} employs transient global attention, inspired by local+global attention from ETC \cite{ainslie2020etc} and integrates summarization-specific pre-training from PEGASUS into the T5 model to handle longer sequences. We use LongT5-base which can handle flexible lengths (unless constrained by memory) due to relative positional embeddings, unlike BART architecture with absolute position embeddings.
\vspace{1mm}

\noindent \textbf{SLED} \cite{ivgi2023efficient} processes long sequences by partitioning them into overlapping chunks and encoding each chunk with a short-range pre-trained encoder. Information across chunks is fused by the decoder by attending to all input tokens, akin to fusion-in-decoder \cite{izacard2021leveraging}. SLED can be applied on top of any short-range model, resulting in SLED-T5 and SLED-LexT5 derived from their respective base models. While it can handle any input length, it is ultimately memory-bound.
\vspace{1mm}

\noindent \textbf{Unlimiformer} \cite{bertsch2023unlimiformer} utilizes a retrieval-based approach to enable short-range pre-trained models to process inputs of unbounded length. It adopts a strategy akin to SLED but focuses solely on the top-k tokens retrieved from a k-nearest-neighbor index constructed over the hidden states of all input tokens at each standard cross-attention head in every decoder layer. This distinguishes Unlimiformer from SLED which is limited by memory when attending to all input tokens in the decoder. We derive Unlimiformer-T5 and Unlimiformer-LexT5 from their base models.
\vspace{1mm}

\noindent \textbf{Evaluation Metrics: } We use ROUGE-1/2/L \cite{lin2004rouge} and BERTScore \cite{zhang2019bertscore} to measure the lexical and semantic overlap between the model generated output and the reference summary. 

\subsection{Results}
We report the results 
across eight LexSumm tasks 
in Table \ref{results}. Notably, the LED consistently outperforms PRIMERA, a difference largely attributed to the contrasting input lengths (16k vs. 4k), particularly evident in R-L scores of datasets with longer inputs like EurLexSumm and UKAbs. Despite PRIMERA's initialization with LED and continued pre-training using the Entity Pyramid masking strategy, we can also attribute its decline to PRIMERA's entity-centric masking strategy 
which turns out to be less suitable for legal corpora. 
This underscores the need for domain-specific masking strategies to facilitate effective transfer. LongT5 demonstrates superior performance compared to PRIMERA and comparable/lower performance to LED, benefiting from its end-to-end pre-training for longer sequences using the Gap Sentence Selection masking. This emphasizes the critical role of 
longer length pre-training 
unlike LED which is not explicitly pre-trained for longer sequences.

\begin{table*}[]
\centering
\scalebox{0.73}{
\begin{tabular}{lcccc}
 \hline
 \multicolumn{1}{c}{\textbf{}} & \textbf{R-1 / 2 / L / BS} & \textbf{R-1 / 2 / L / BS} & \multicolumn{1}{c}{\textbf{R-1 / 2 / L / BS}} & \textbf{R-1 / 2 / L / BS} \\ \hline
\multicolumn{1}{c}{\textbf{}} & \textbf{BillSum}          & \textbf{EurLexSumm}       & \textbf{GovReport}  & \textbf{MLS-Long}         \\ \hline
GPT-3.5-Turbo & 31.0 / 13.3 / 27.9 / 61.9 & 22.1 / 6.9 / 19.4 / 62.0 & 24.4 / 8.1 / 22.0 / 60.2 & 24.2 / 8.7 / 21.8 / 59.9 \\
Claude Instant & 31.5 / 13.5 / 28.5 / 61.5 & 24.0 / 8.2 / 21.9 / 61.9 & 28.5 / 8.8 / 26.1 / 61.4 & 29.1 / 10.8 / 26.6 / 61.2 \\
\textcolor{gray!70}{Unlim-LexT5} & \textcolor{gray!70}{37.1 / 21.8 / 33.9 / 65.8 } & \textcolor{gray!70}{34.8 / 17.7 / 30.1 / 66.6 }& \textcolor{gray!70}{37.2 / 17.3 / 34.4 / 64.9} &  \textcolor{gray!70}{ 37.9 / 17.2 / 34.8 / 67.1}\\
\hline                                    
\multicolumn{1}{c}{\textbf{}} & \textbf{MLS-Short}        & \textbf{MLS-Tiny}         & \textbf{INAbs}            & \textbf{UKAbs}            \\ \hline
GPT-3.5-Turbo & 21.8 / 7.95 / 19.5 / 56.9 & 15.3 / 3.3 / 12.8 / 49.3 & 20.8 / 6.6 / 18.3 / 58.1 & 24.2 / 7.8 / 21.6 / 59.0 \\
Claude Instant & 27.7 / 10.3 / 25.6 / 57.8 & 16.5 / 3.4 / 13.5 / 50.2 & 23.9 / 7.8 / 21.8 / 60.3 & 29.0 / 9.6 / 26.6 / 61.9\\
\textcolor{gray!70}{Unlim-LexT5} & \textcolor{gray!70}{35.2 / 17.8 / 33.4/ 64.8} & \textcolor{gray!70}{26.6 / 11.8 / 22.6 / 56.2} & \textcolor{gray!70}{36.5 / 16.6 / 32.1 / 63.1} & \textcolor{gray!70}{34.8 / 14.2 / 31.3 / 64.3} \\
\hline
\end{tabular}}
\caption{Evaluation results of LLM models across eight subsampled test datasets of LexSumm.}
\label{zero_results}
\end{table*}

SLED-T5 and Unlimiformer-T5 exhibit comparable performance to long-range pre-trained models like LED and LongT5, even surpassing PRIMERA in most datasets. This suggests that leveraging off-the-shelf short-range pre-trained language models and integrating them into frameworks for longer context tasks can yield competitive results. Our LexT5 models, pre-trained on legal corpora using random span masking strategies without specific long-range or summarization pre-training, when plugged into SLED and Unlimiformer consistently outperform all others across all datasets, particularly excelling in more challenging higher n-gram metrics (R-2, R-L). This underscores the importance of domain-specific training and thanks to the flexibility of these frameworks that allow easy integration of any pre-trained short-range language models without the need for expensive long-sequence pre-training. Furthermore, Unlimiformer-LexT5 outperforms SLED-LexT5 in 7 out of 8 datasets, indicating that attending only to the top-k input keys can be an accurate approximation of full attention, motivating design of effective retrieval methods to handle long context processing. 

\vspace{1mm}
\noindent \textbf{Zero-shot evaluation with LLMs:} We use stratified sampling to select 50 instances from each of test split of the LexSumm dataset, across diverse input lengths. We always include the most 10 longest inputs from test set and sampled 10, 15, 15 from the three buckets derived from rest of the test set based on their input lengths. We evaluate two long-context based LLM models - Claude-Instant-1.2 and GPT-3.5-Turbo with hierarchical merging strategy for summarization following \citet{chang2023booookscore} where in the input is divided it into smaller chunks to summarize individually and then partial summaries are repeatedly merged to form final summary. Detailed illustration and prompts are in App. \ref{zero_prompts}. We reported the performance of these models in Table \ref{zero_results}. We observe that Claude model performing better than GPT-3.5-Turbo across all the datasets consistently. On comparing with fine-tuned variant of Unlimiformer-LexT5, we observe fine-tuned variant performing better compared to them, in most challenging ROUGE-2 and -L scores.

\vspace{1mm}
\noindent \textbf{Qualitative Analysis:} We examine outputs from PRIMERA and LED on the In-Abs case in \ref{output_len}. PRIMERA's summary completely misrepresents the case by incorrectly stating that the issue concerns the validity of dismissal orders under "r. 149 of the Code of Civil Procedure," whereas it should refer to Rules 148(3) and 149(3) of the Indian Railway Establishment Code, focusing on whether they violate articles 14 and 311(2) of the Constitution of India. The summary's focus omits details about the Supreme Court’s decision. Although the phrase "code of civil procedure" is mentioned in the input, it is unrelated to the context in the summary. PRIMERA's summary emphasizes procedural details, while the original text primarily discusses procedural fairness under article 311(2). This discrepancy in understanding the case's context and focus of the summary is attributed to the limited input context of PRIMERA. While the 16k-based LED attempts to produce a more faithful summary, it reduces a multi-applicant case to a single one and incorrectly mentions "under Rule 148" instead of the specific Rules 148(3) and 149(3), resulting in misrepresentation. LED still struggles to accurately capture the final outcome presented towards the end of the 39k-token input. To analyze the impact of legal pre-training, we compare Unlimiformer-T5 with LexT5 using GovReport input on climate change in App. \ref{output_lex}. While the T5 introduces Government Accountability Office (GAO) in summary, not even mentioned in the input, LexT5 avoids such entity-level hallucinations but emphasizes only on certain portions such as the U.S. climate policy landscape, leaving discussion on pitfalls.

We analyze outputs from the MLS-Tiny dataset, tackling a needle-in-the-haystack problem to distill crucial case details into a single tweet-like sentence. Reference summary and various model generations are presented in App \ref{MLS-Tiny}. The document outlines a legal complaint by the American-Arab Anti-Discrimination Committee against U.S. Customs and Border Protection, alleging wrongful withholding of records. These records pertain to Arab and Muslim American residents being unfairly removed from the Global Entry program. The conclusion indicates a consensus that previously secret records will be disclosed. PRIMERA captures the essence but omits the legal basis (FOIA) mentioned in the reference summary. Its resemblance to a full sentence rather than a Twitter post style can be attributed to its pre-training objective of gap sentence generation, making it less adaptable to switch to a Twitter style. LED summary highlights the action succinctly but generalizes it to a travel ban. LongT5 misses and misrepresents main information, being partially unfaithful. SLED and Unlimiformer summaries partially present the lawsuit but omit resolution details, indicating the challenge of fusing information across chunks. Lex summaries provide additional details but struggle to synthesize final outcome into the summary.

We present the zero-shot outputs from GPT-3.5 and Claude on the IN-Abs in App. \ref{zero-shot-summaries}. Both summaries offer a high-level abstraction of the case details, focusing on the main legal issue under scrutiny and the court's findings. Despite differing from the reference summary style, both summaries effectively highlight key document aspects, ensuring easy understanding, albeit with some pertinent details omitted. Claude provides more complete and grounded summary than GPT-3.5 by elaborating on crucial elements like Article 311(2) of the Constitution. Future work should assess the quality of these generations on large scale with diverse legal experts given the subjective nature of quality.

\section{Conclusion}
In this work, we curate LexSumm benchmark for training and evaluating legal summarization tasks in English. LexSumm can serve as an evaluation platform to foster development of approaches dealing with long legal text using efficient transformer architectures or retrieval-based methods adopted for longer context, legal-oriented pre-training or masking schemes, faithful decoding strategies. We pre-train LexT5, a legal seq2seq model and evaluate on LegalLAMA probing task and LexSumm downstream benchmark. We compare LexT5 wrapped in long-range adaptation frameworks such as SLED and Unlimiformer with T5 model in long-range adaptation, other long-range pre-trained models, and even zero-shot LLMs. We release LexT5 to the community, hoping it will serve as a backbone model for various legal generative tasks. Additionally, we envision LexSumm evolving into a dynamic benchmark, expanding with new datasets over time.

\section*{Limitations}
An important limitation of our benchmark is its reliance on English-only evaluation, which limits the generalizability of our findings to legal systems operating in languages other than English. Given the global nature of legal systems, each conducting proceedings in their official languages, there is a clear need for multilingual legal generative models. However, our ability to develop such models is hindered by the scarcity of multilingual legal generative task data, except for Chinese datasets. Furthermore, our dataset predominantly consists of data from English-speaking nations, where data availability is more accessible, thereby constraining the diversity and inclusivity of our study. Overcoming this limitation poses additional challenges, including bureaucratic hurdles in accessing court records, dependence on outdated technology for managing legal documents and privacy concerns related to contracts. Additionally, obtaining annotated data for downstream tasks proves to be expensive due to the need for specialized legal expertise.

Our LexSumm evaluation primarily relies on established summarization metrics such as ROUGE and BERTScore. While these metrics have been used in many prior works on legal document summarization and are known to provide a quantitative measure of summarization quality, they may not fully capture the nuanced legal content, context and intricacies essential for legal professionals. A potential avenue for further research could be developing additional legal domain-specific evaluation metrics. Another significant limitation of our study is the absence of direct participation or validation by legal experts in the assessment of summarization outputs, which we could not perform due to lack of access to legal experts.

Although LexT5 has primarily been evaluated on summarization tasks within LexSumm, we intend to broaden its evaluation scope to include Legal NLU and other generation tasks such as simplification or translation. Evaluating seq2seq models on Legal NLU datasets like LexGLUE \cite{chalkidis2022lexglue} poses a challenge due to the multi-label nature of tasks. This complexity necessitates additional modifications to enable seq2seq models for multi-label tasks \cite{kementchedjhieva2023exploration}.

\section*{Ethics Statement}
All datasets incorporated into LexSumm are openly accessible and have been previously published, with citations provided to the original sources. We strongly encourage users of LexSumm to acknowledge these sources, suggesting referencing this work alongside citing the original sources when utilizing multiple LexSumm datasets and employing the LexSumm evaluation framework. Otherwise, citation of only the original sources is appropriate.

The aim of LexSumm is to introduce a unified legal NLP benchmark to expedite the development of legal models and assess various technical approaches in handling legal tasks. By offering a comprehensive benchmark spanning multiple jurisdictions, this initiative aims to provide guidance to system developers on best practices, serve as a crucial yardstick for measuring progress and guide research efforts, ultimately aiding practitioners in creating supportive technology tailored for legal professionals and laypersons alike.

While datasets in LexSumm such as EurLexSumm, BillSum, and GovReport primarily consist of legislation or policy material and are unlikely to contain personal data, other datasets like MultiLexSum, UKAbs, and InAbs contain personal data of the parties and individuals involved in legal proceedings. However, these datasets are published by respective courts in accordance with data protection laws. We do not anticipate any harm resulting from our experiments beyond the disclosure of this information.

We train and release the LexT5 model using historical legal data sourced from prior work on LeXFiles \cite{chalkidis-etal-2023-lexfiles}. These historical corpora inherently encode biases and inequities present within the legal domain, which might be inherited by these models. Deploying LexT5 without robust scrutiny and mitigation strategies could perpetuate and amplify these biases, potentially leading to unjust outcomes in legal decision-making processes. Furthermore, the widespread adoption of LexT5 in legal applications could exacerbate disparities in access to justice, as marginalized communities may be disproportionately affected by biased model predictions. To address these ethical concerns, it is imperative to conduct thorough bias audits, implement mitigation techniques, ensure transparency and accountability in model deployment, and continuously monitor and evaluate the model's performance in real-world settings.

Moreover, fine-tuned models developed for each specific task of LexSumm may exhibit performance variations across different partitions within the same legal domain. For instance, as highlighted in \citealt{agarwal2022extractive}. in contexts like the Board of Veterans’ Appeals, cases involving rarely occurring disabilities or specialized legal and military situations may lead to suboptimal summaries due to sparsity in the training data. This variability could disproportionately impact groups that should be treated equally if their characteristics coincide with these less frequent legal configurations. Engaging domain experts to curate datasets with better representation across different types of injuries and legal phenomena can be a proactive step in enhancing the model’s understanding of uncommon or group-related legal contexts, potentially mitigating disparities in performance.

\bibliography{custom}
\bibliographystyle{acl_natbib}

\appendix

\section{Data Characteristics}
\begin{figure*}[!ht]
    \centering
    \begin{subfigure}{0.3\textwidth}
        \centering
        \includegraphics[width=\linewidth]{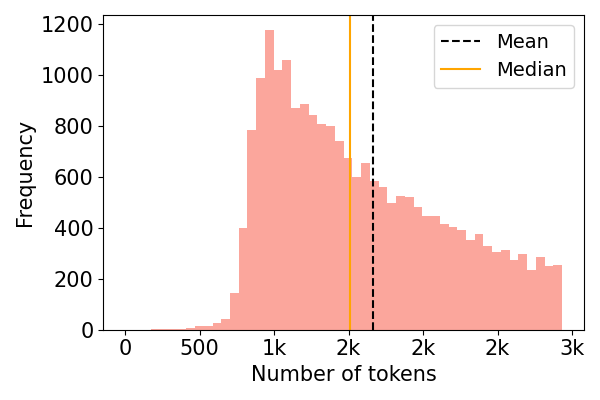}
        \caption{Input length distr. - BillSum}
    \end{subfigure}
    \hfill
    \begin{subfigure}{0.3\textwidth}
        \centering
        \includegraphics[width=\linewidth]{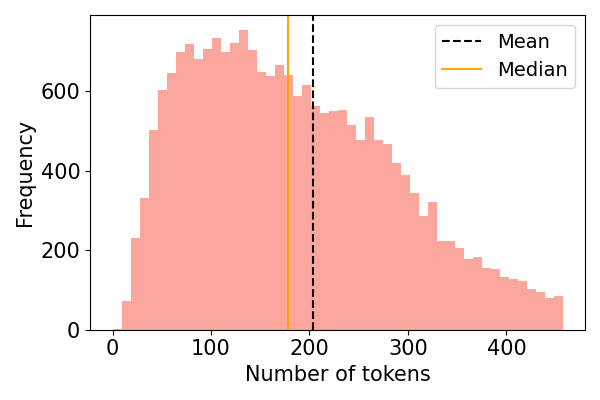}
        \caption{Summary length distr. - BillSum}
    \end{subfigure}
    \hfill
    \begin{subfigure}{0.3\textwidth}
        \centering
        \includegraphics[width=\linewidth]{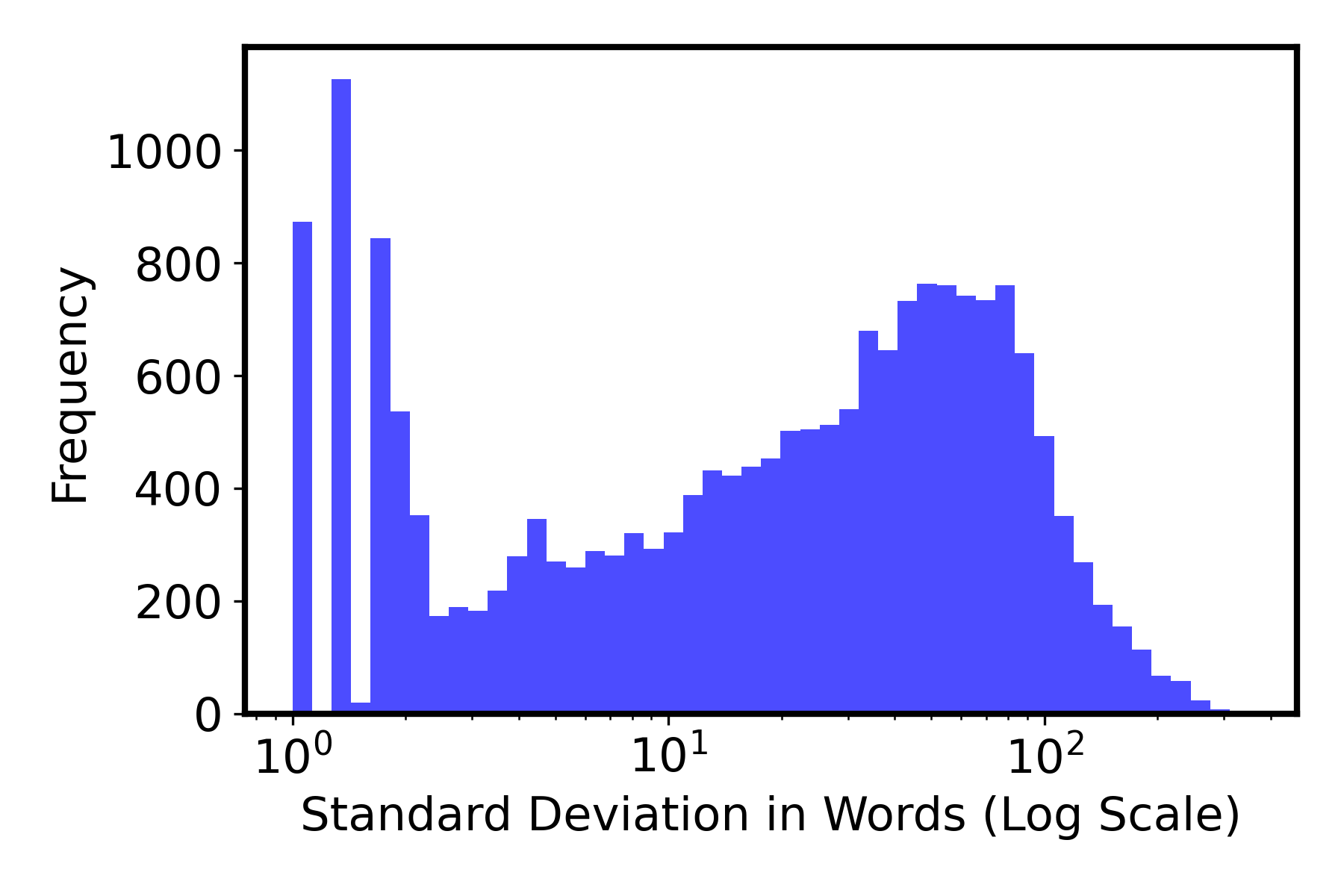}
        \caption{Fusion Score distr. - BillSum}
    \end{subfigure}
    \vfill
    \begin{subfigure}{0.3\textwidth}
        \centering
        \includegraphics[width=\linewidth]{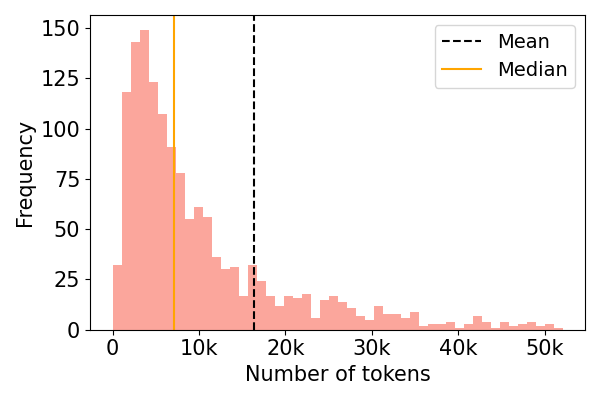}
        \caption{Input length distr. - EurLexSum}
    \end{subfigure}
    \hfill
    \begin{subfigure}{0.3\textwidth}
        \centering
        \includegraphics[width=\linewidth]{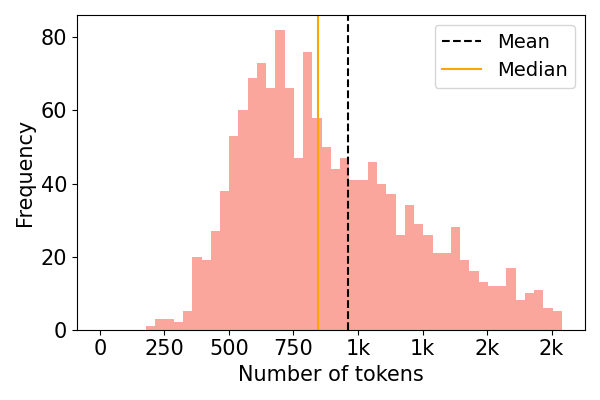}
        \caption{Summary len. distr. - EurLexSum}
    \end{subfigure}
    \hfill
    \begin{subfigure}{0.3\textwidth}
        \centering
        \includegraphics[width=\linewidth]{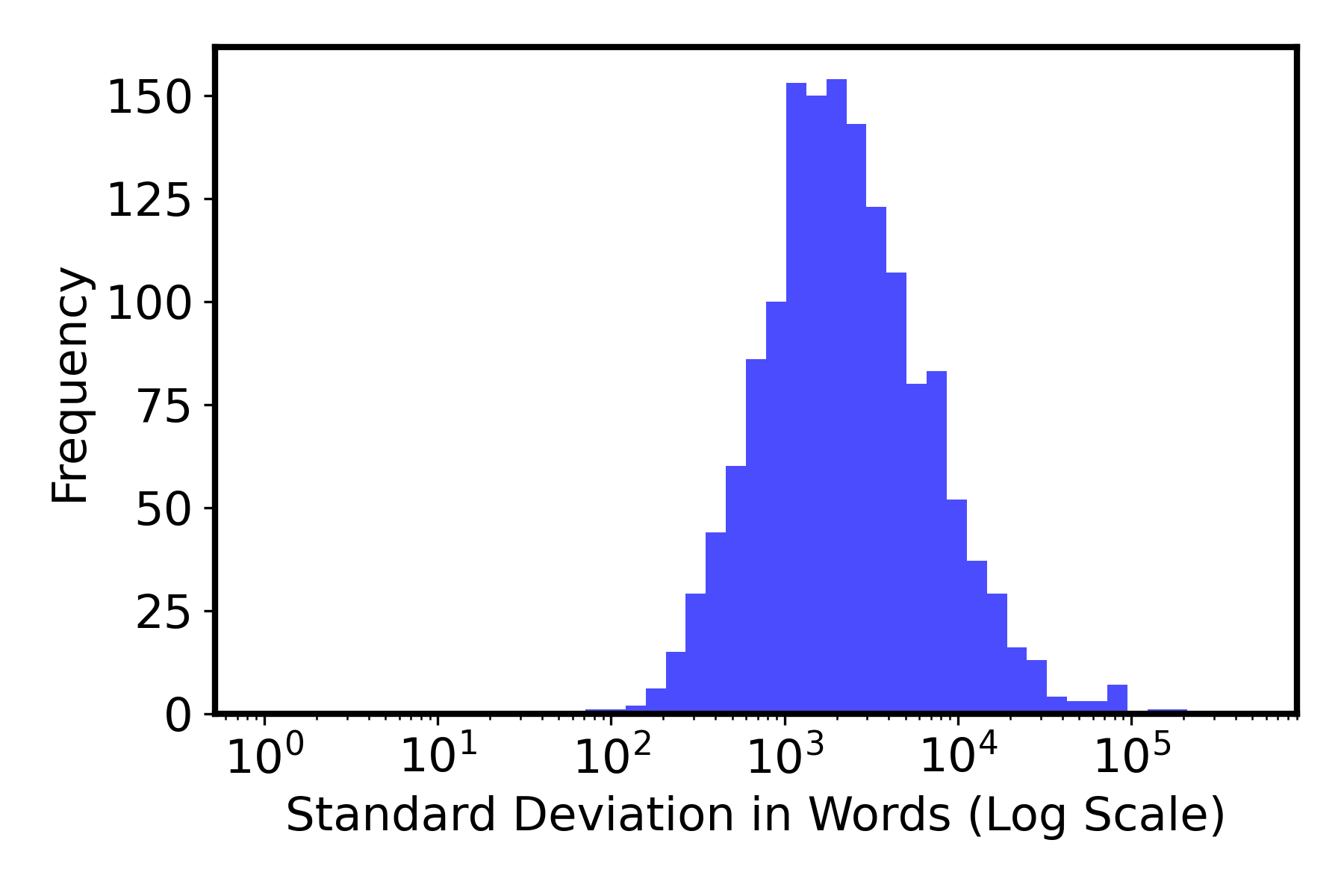}
        \caption{Fusion Score distr. - EurLexSum}
    \end{subfigure}
    \vfill
    \begin{subfigure}{0.3\textwidth}
        \centering
        \includegraphics[width=\linewidth]{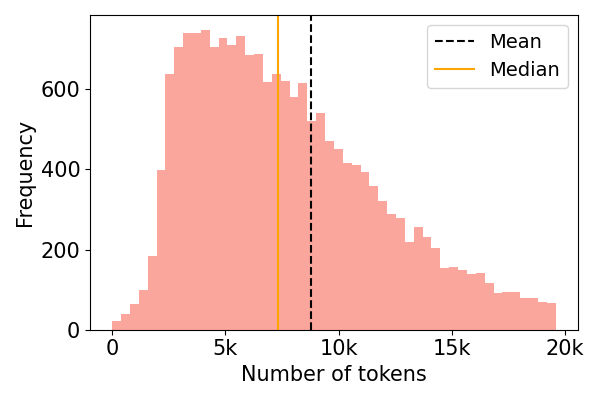}
        \caption{Input length distr. - GovReport}
    \end{subfigure}
    \hfill
    \begin{subfigure}{0.3\textwidth}
        \centering
        \includegraphics[width=\linewidth]{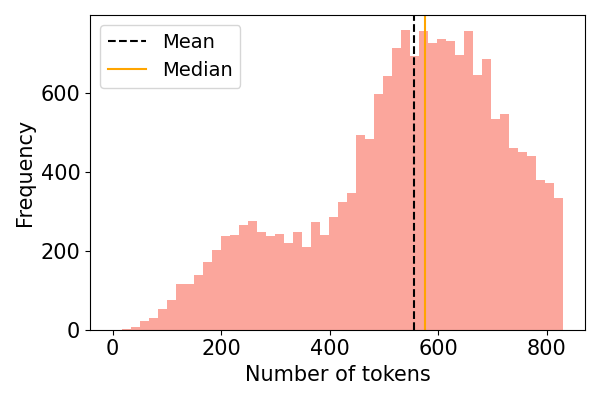}
        \caption{Summary length distr. - GovReport}
    \end{subfigure}
    \hfill
    \begin{subfigure}{0.3\textwidth}
        \centering
        \includegraphics[width=\linewidth]{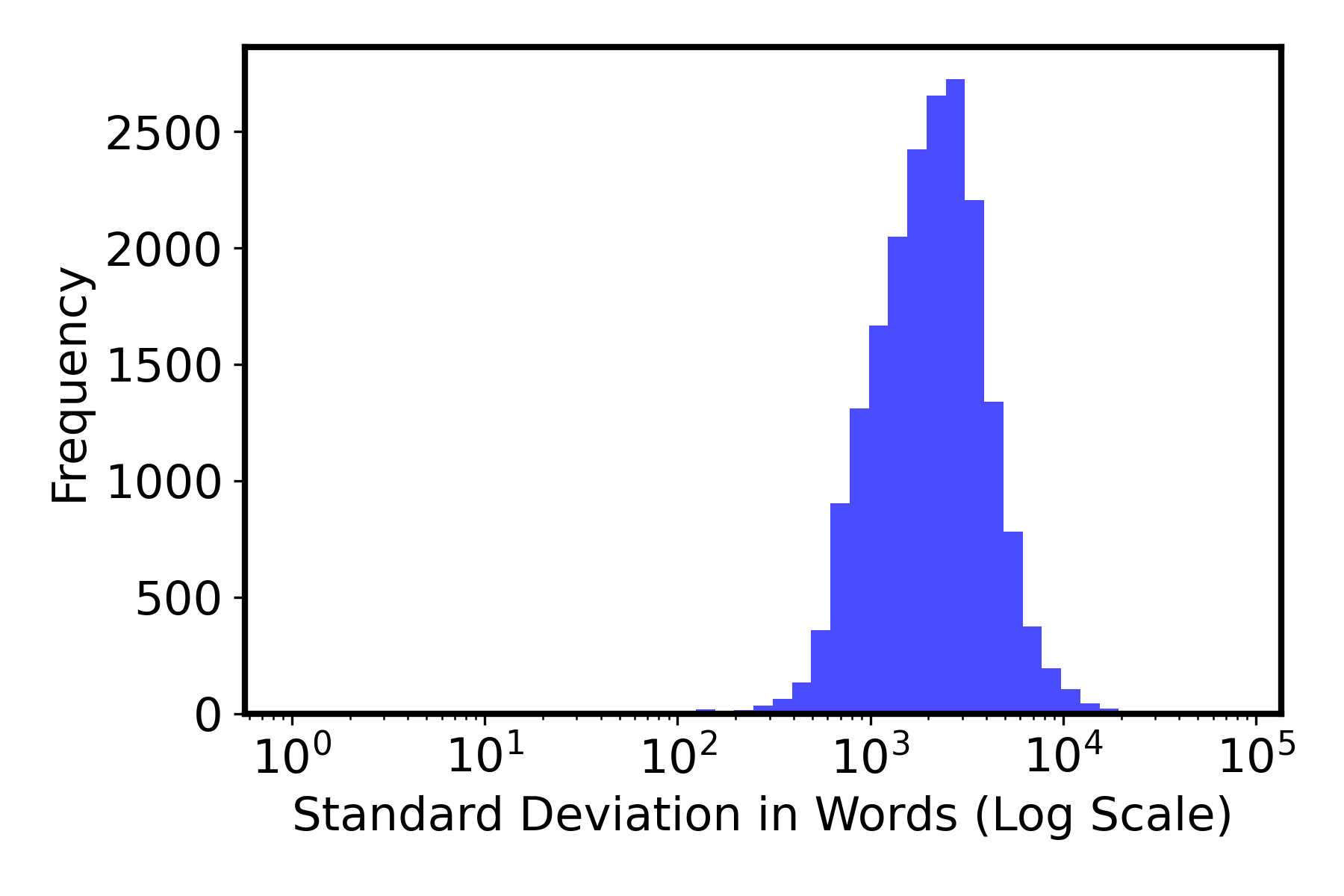}
        \caption{Fusion Score distr. - GovReport}
    \end{subfigure}
    \vfill
    \begin{subfigure}{0.3\textwidth}
        \centering
        \includegraphics[width=\linewidth]{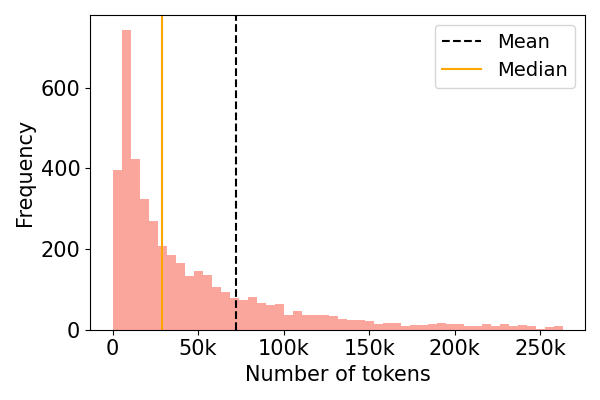}
        \caption{Input length distr. - MLS-Long}
    \end{subfigure}
    \hfill
    \begin{subfigure}{0.3\textwidth}
        \centering
        \includegraphics[width=\linewidth]{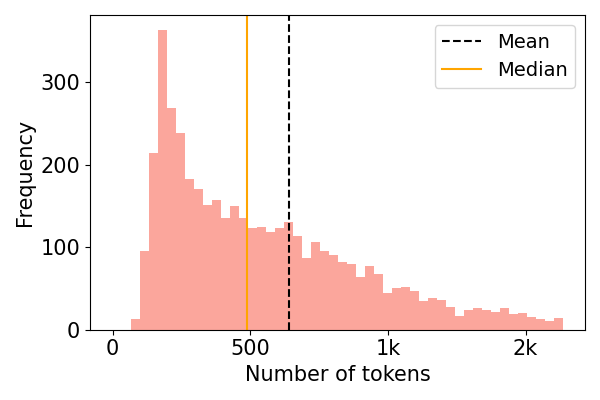}
        \caption{Summary length distr. - MLS-Long}
    \end{subfigure}
    \hfill
    \begin{subfigure}{0.3\textwidth}
        \centering
        \includegraphics[width=\linewidth]{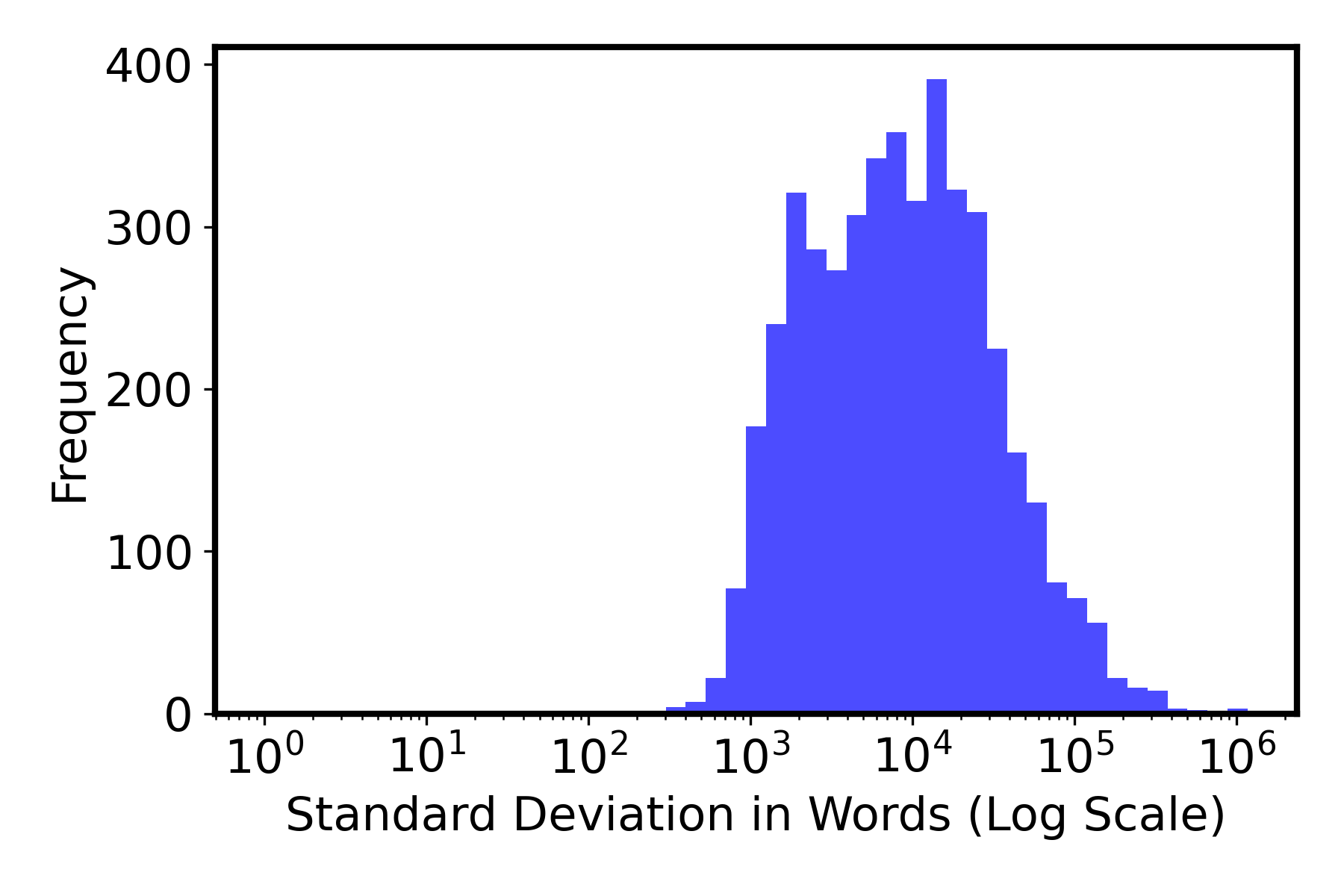}
        \caption{Fusion Score distr. - MLS-Long}
    \end{subfigure}
    \vfill
    \begin{subfigure}{0.3\textwidth}
        \centering
        \includegraphics[width=\linewidth]{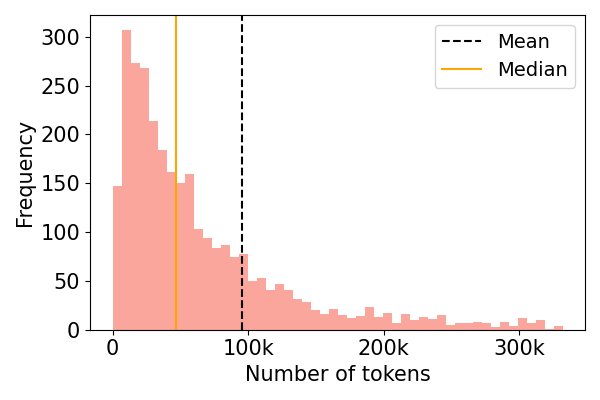}
        \caption{Input length distr. - MLS-Short}
    \end{subfigure}
    \hfill
    \begin{subfigure}{0.3\textwidth}
        \centering
        \includegraphics[width=\linewidth]{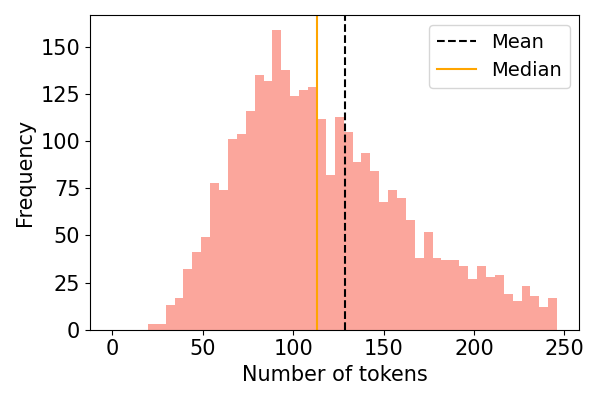}
        \caption{Summary length distr. - MLS-Short}
    \end{subfigure}
    \hfill
    \begin{subfigure}{0.3\textwidth}
        \centering
        \includegraphics[width=\linewidth]{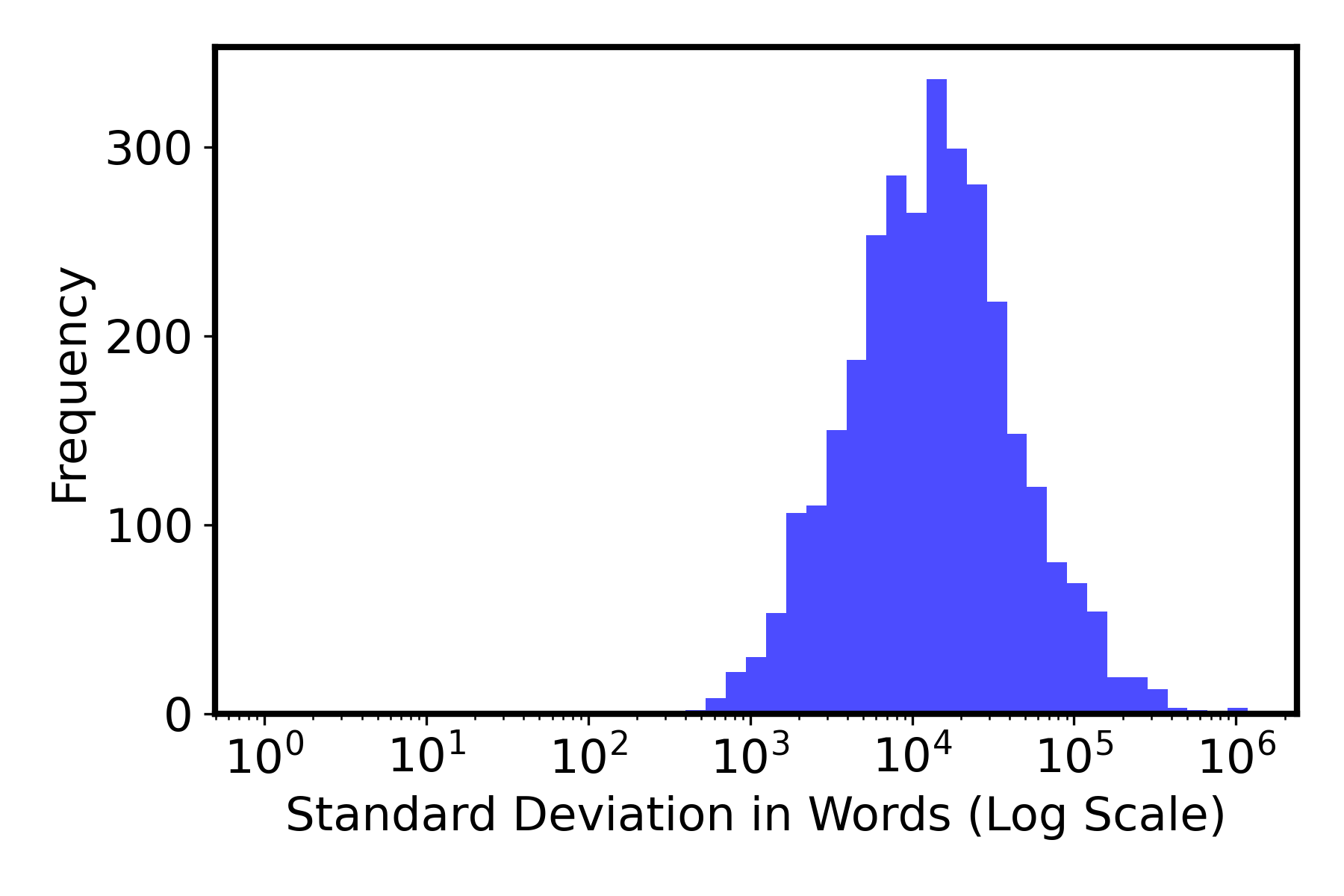}
        \caption{Fusion Score distr. - MLS-Short}
    \end{subfigure}
    \caption{Distribution of input length, summary length and fusion scores for LexSumm datasets.}
    \label{fig:data-char1}
\end{figure*}

\begin{figure*}[!ht]
    \centering
    \begin{subfigure}{0.3\textwidth}
        \centering
        \includegraphics[width=\linewidth]{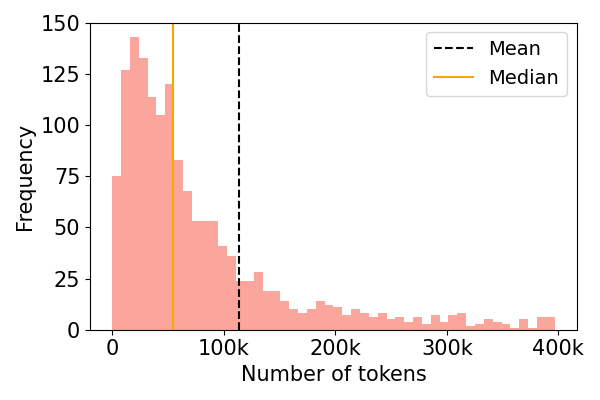}
        \caption{Input length distr. - MLS-Tiny}
    \end{subfigure}
    \hfill
    \begin{subfigure}{0.3\textwidth}
        \centering
        \includegraphics[width=\linewidth]{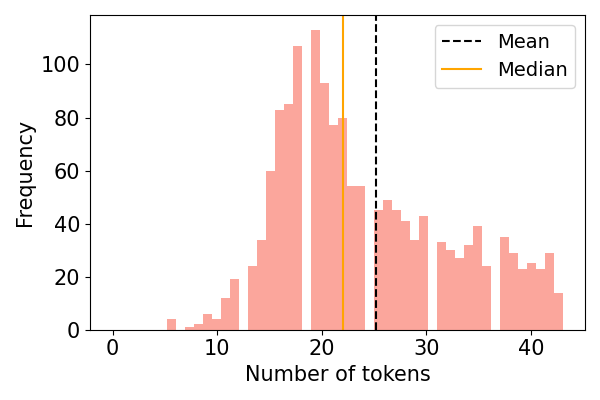}
        \caption{Summary length distr. - MLS-Tiny}
    \end{subfigure}
    \hfill
    \begin{subfigure}{0.3\textwidth}
        \centering
        \includegraphics[width=\linewidth]{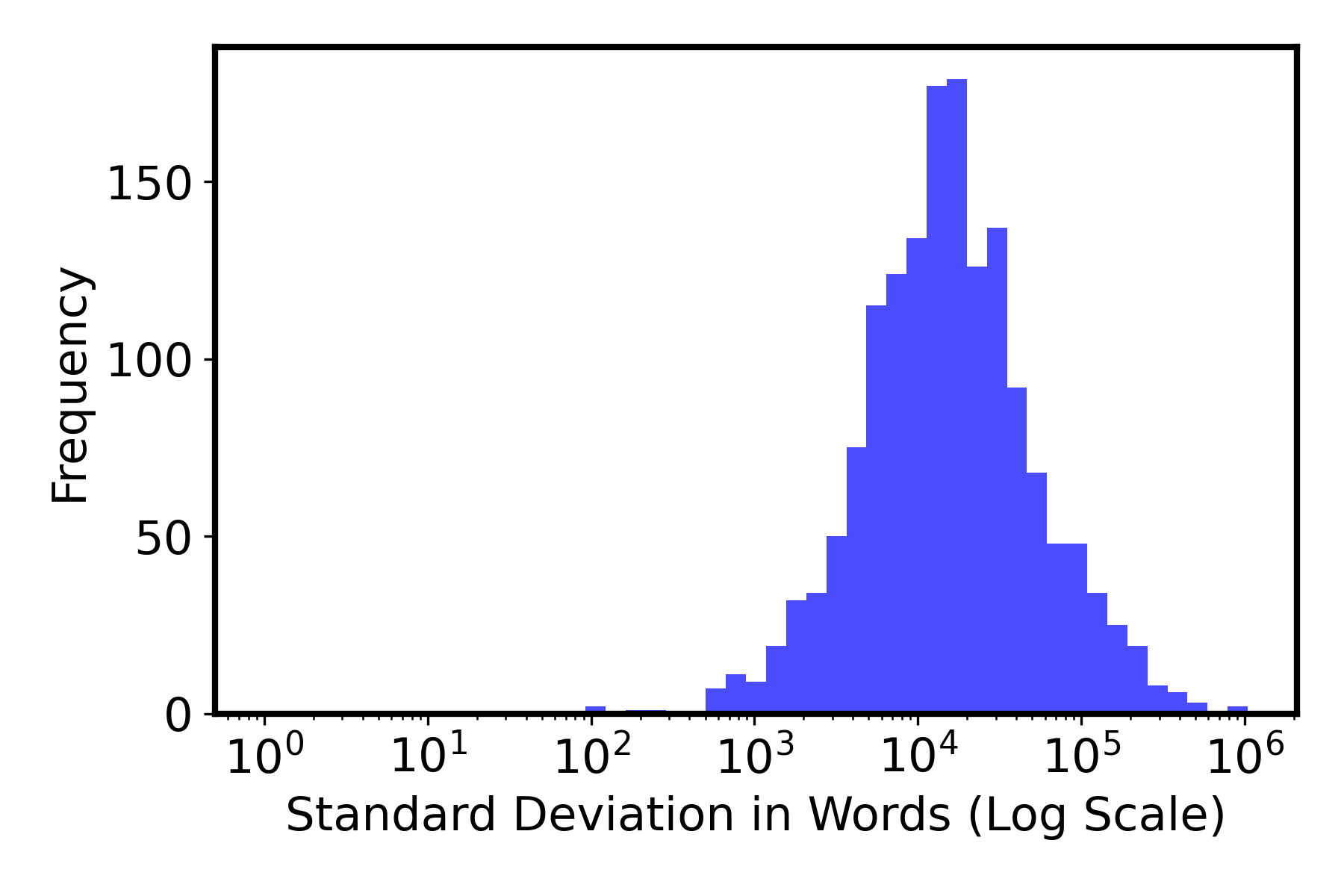}
        \caption{Fusion Score distr. - MLS-Tiny}
    \end{subfigure}
    \vfill
    \begin{subfigure}{0.3\textwidth}
        \centering
        \includegraphics[width=\linewidth]{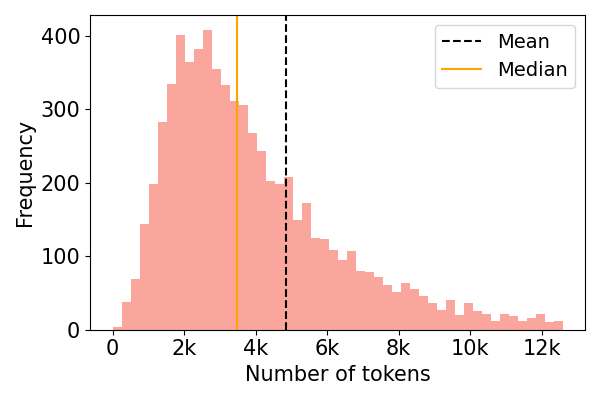}
        \caption{Input length distr. - InAbs}
    \end{subfigure}
    \hfill
    \begin{subfigure}{0.3\textwidth}
        \centering
        \includegraphics[width=\linewidth]{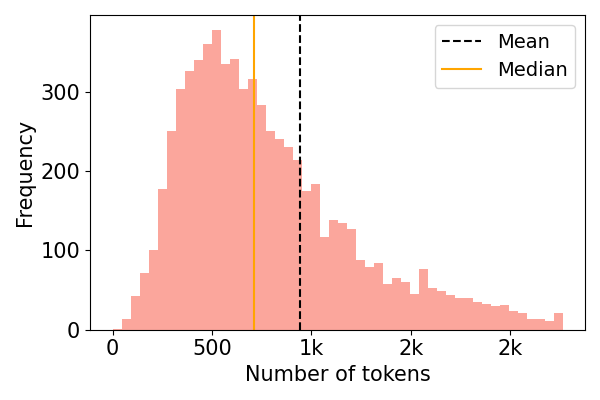}
        \caption{Summary length distr. - InAbs}
    \end{subfigure}
    \hfill
    \begin{subfigure}{0.3\textwidth}
        \centering
        \includegraphics[width=\linewidth]{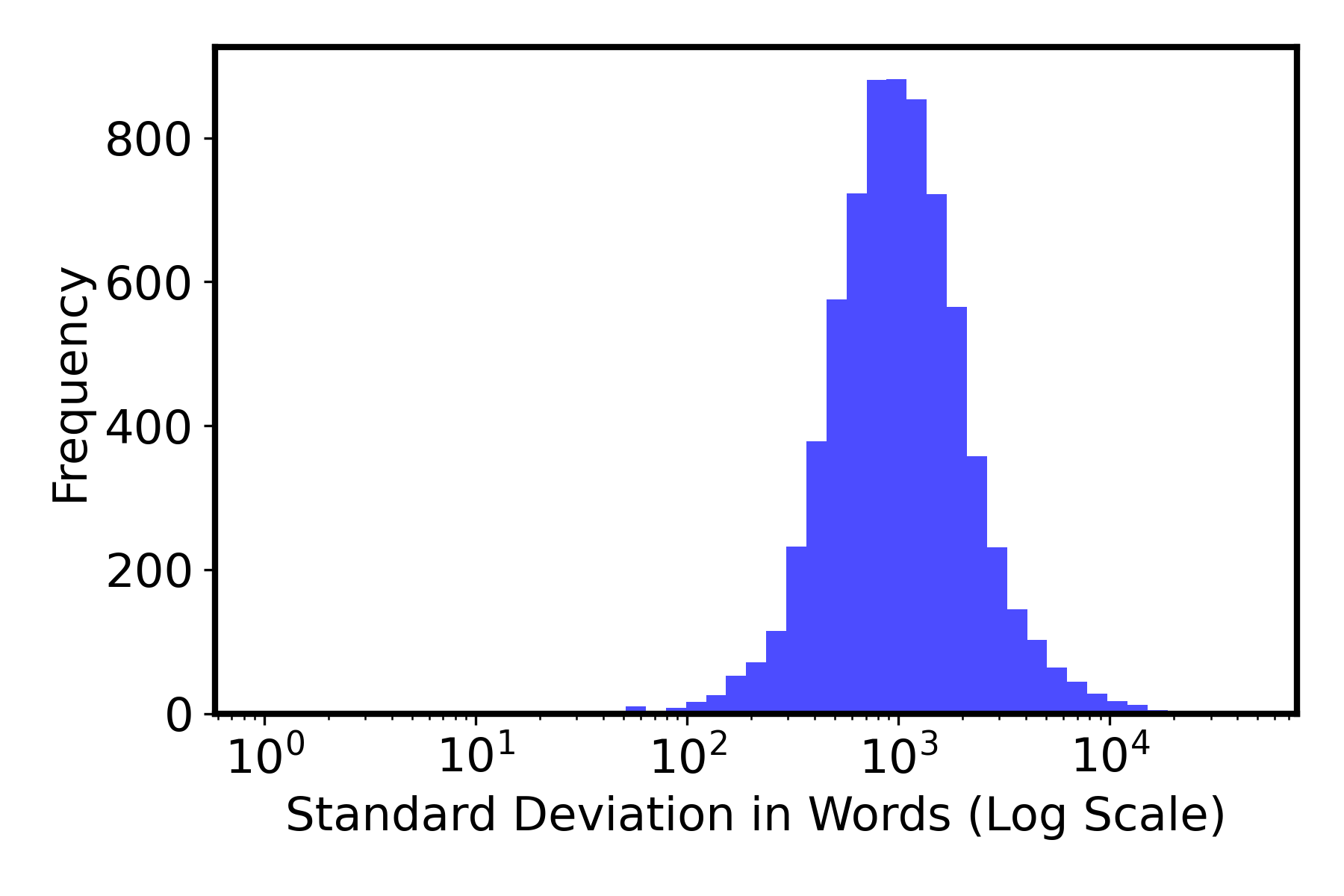}
        \caption{Fusion Score distr. - InAbs}
    \end{subfigure}
    \vfill
    \begin{subfigure}{0.3\textwidth}
        \centering
        \includegraphics[width=\linewidth]{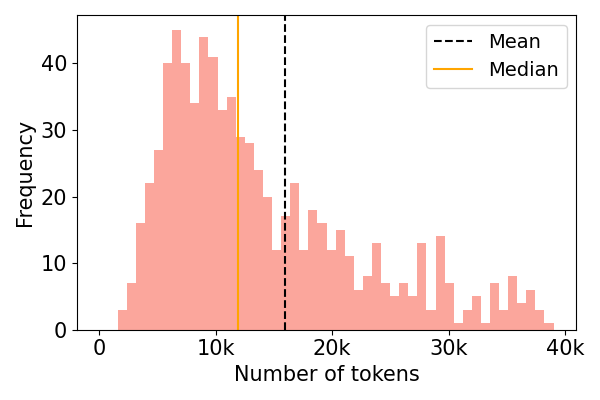}
        \caption{Input length distr. - UKAbs}
    \end{subfigure}
    \hfill
    \begin{subfigure}{0.3\textwidth}
        \centering
        \includegraphics[width=\linewidth]{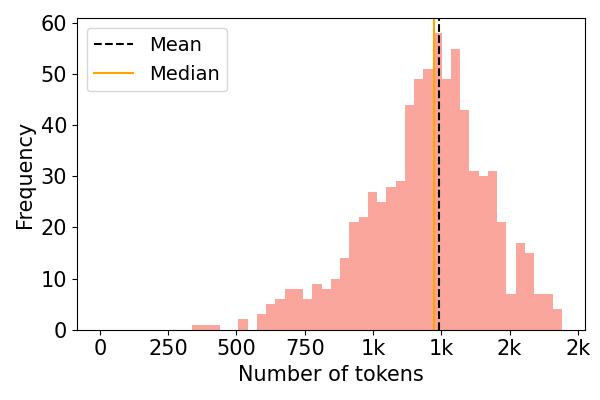}
        \caption{Summary length distr. - UKAbs}
    \end{subfigure}
    \hfill
    \begin{subfigure}{0.3\textwidth}
        \centering
        \includegraphics[width=\linewidth]{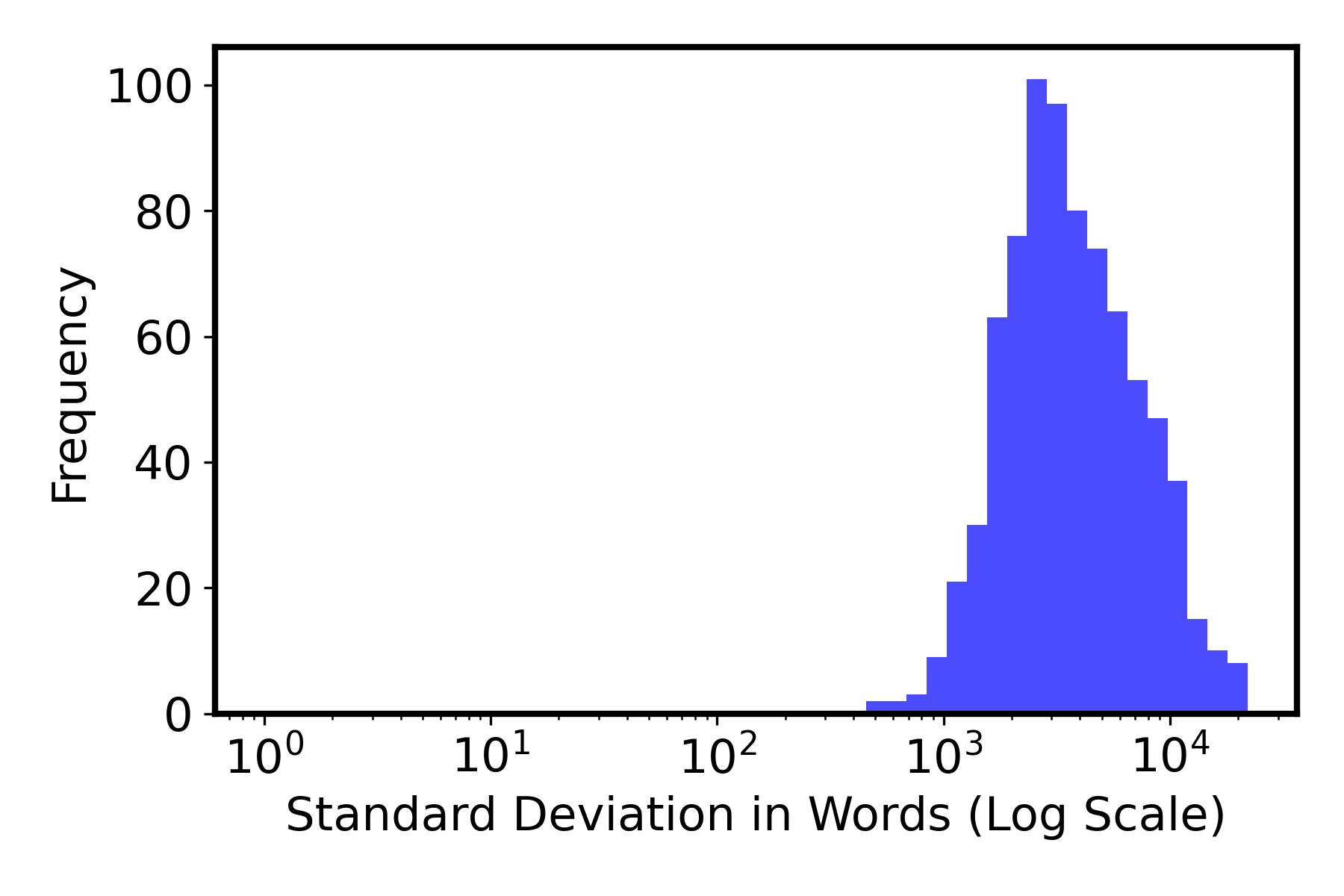}
        \caption{Fusion Score distr. - UKAbs}
    \end{subfigure}
    \caption{Distribution of input length, summary length and fusion scores for LexSumm datasets.}
    \label{fig:data-char2}
\end{figure*}

Fig. \ref{fig:data-char1} and \ref{fig:data-char2} display the input text length, summary text length and fusion score distribution for each of the dataset in LexSumm benchmark.

\section{Implementation Details for LexT5}
\label{pre-impl}
We use a learning rate of 0.005, linear warmup of 2.5k steps, inverse square root learning rate decay, maximum sequence length of 512 and is pre-trained for 250k steps. 
We employ a batch size of 65536 tokens and is optimized end-to-end using Adafactor optimizer \cite{shazeer2018adafactor} with a corrupted token ratio of 15\% with the mean noise span length of 3. Pre-training is carried out using  Google Cloud TPU with 8 cores (v3.8). 

\section{Implementation Details for downstream tasks}
\label{impl-details}
We fine-tune each of our models on individual datasets using the AdamW optimizer \cite{loshchilov2018decoupled} with hyperparameters $\beta = (0.9, 0.98)$ and $\epsilon = 1e-6$, alongside mixed precision (fp16) and gradient checkpointing techniques. For consistency, we set the maximum target sequence length to 512 across all models, while the input sequence length is set to 16384 for all models except PRIMERA and LongT5, which support 4096 and 8192 tokens, respectively, during training. We train LongT5, PRIMERA, and LongT5 with a learning rate of 2e-5, while Unlimiformer and SLED are trained with a learning rate of 1e-4 for 15 epochs. To control the learning rate, we employ a scheduler that warms up from zero during the first 10\% of the steps and then linearly decays back to zero for the remaining steps. For models utilizing chunking, we set the chunk overlap ratio to 0.5. During inference, we set the minimum length to 16 for datasets with shorter outputs such as BillSum, MultiLexSumm-Tiny, and MultiLexSumm-Short, and to 128 for the remaining datasets. The maximum length is set to 16384 to ensure the model generates text without abruptly ending. Additionally, we utilize four beams for datasets with longer outputs and seven beams for datasets with shorter outputs. We apply a length penalty of 0.8 and 2 for datasets with shorter and longer outputs, respectively. Early stopping is disabled for datasets with longer outputs and enabled for datasets with shorter outputs.

\section{Zero-shot Summarization}
\label{zero_prompts}
\begin{figure*}
    \centering
    \includegraphics[width=0.8\textwidth]{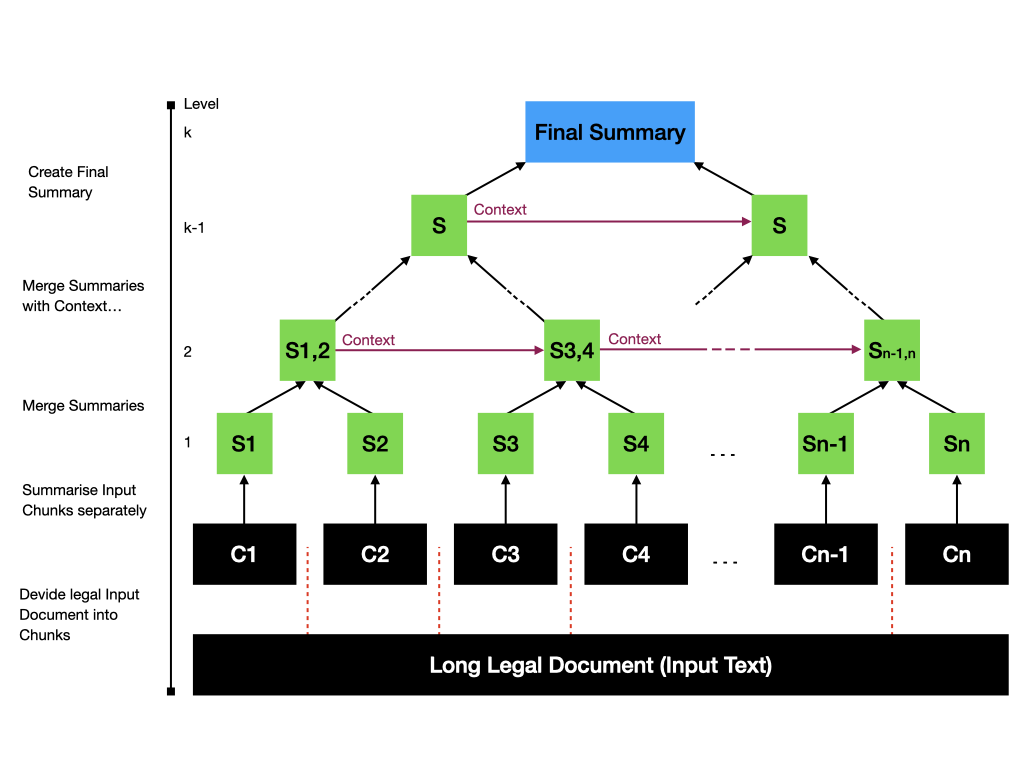}
    \caption{Visualization of Hierarchical merging strategy for summarization of Long Inputs: "S" represents the summary, "C" denotes the chunk, "n" is the total number of chunks, and "k" indicates the maximum number of levels}
    \label{fig:hier}
\end{figure*}

An illustration of hierarchical merging strategy for long input summarization can be visualized in Fig. \ref{fig:hier}. Hierarchical merging strategy requires three prompts as follows:

\noindent \textbf{(i) Summarizing an input chunk:}

\noindent \texttt{Below is a part of a legal document:}

\noindent \texttt{---}

\noindent \texttt{\{input\}}

\noindent \texttt{---}

\noindent \texttt{We are creating one comprehensive summary for the legal document by recursively merging summaries of its chunks. Now, write a summary for the excerpt provided above, making sure to include vital information  related to legal arguments, backgrounds, legal settings, key figures, their objectives, and motivations. If a legal norm or code is cited,  it must be correct and include the right number. Summarize all key events and everything that is relevant to the case. Be concise and use legal notation and language. The summary must be within \{words\} and could include multiple paragraphs.}

\vspace{4mm}

\noindent \textbf{(ii) Merging two chunk-level summaries:}

\vspace{2mm}

\noindent \texttt{Below are several summaries of consecutive parts of a legal document:}

\noindent \texttt{---}

\noindent \texttt{\{input\}}

\noindent \texttt{---}

\noindent \texttt{We are creating one comprehensive summary for the legal document by recursively merging summaries of its chunks. Now, merge the given summaries into one single summary, making sure to include vital information related to legal arguments, backgrounds, legal settings, key figures, their objectives, and motivations. The summary must be within {words} and could include multiple paragraphs.}

\vspace{4mm}

\noindent \textbf{(iii) Merging two summaries with added context from previously-generated merged summaries}

\vspace{2mm}

\noindent \texttt{Below is a summary of the context preceding some parts of a legal document:}

\noindent \texttt{---}

\noindent \texttt{\{context\}}

\noindent \texttt{---}

\noindent \texttt{Below are several summaries of consecutive parts of a legal document:}

\noindent \texttt{---}

\noindent \texttt{\{input\}}

\noindent \texttt{---}

\noindent \texttt{We are creating one comprehensive summary for the legal document by recursively merging summaries of its chunks. Now, merge the preceding context and the summaries into one single summary, making sure to include vital information related to legal arguments, backgrounds, legal settings, key figures, their objectives, and motivations. The  summary must be within {words} and could include multiple paragraphs.}

\vspace{4mm}
The prompts above have been used for all datasets in the LexSummZero benchmark, except MLS - Tiny dataset, where the output is a single-sentence Twitter post and the following prompts are used for that dataset.

\vspace{4mm}

\noindent \textbf{(i) Summarizing an input chunk:}

\vspace{2mm}

\noindent \texttt{Below is a part of a legal document:}

\noindent \texttt{---}

\noindent \texttt{\{input\}}

\noindent \texttt{---}

\noindent \texttt{We are creating one comprehensive summary for the legal document, stylized as a single-sentence Twitter post. This summary should encapsulate the most relevant information: who is involved, when did it happen, to whom it concerns, on what legal basis, and the location (as a shortened reference). Ensure to capture key legal arguments, backgrounds, legal settings, key figures, their objectives, and motivations. If a legal norm or code is cited, include the correct number succinctly. Despite the complexity of legal arguments, references to precedent cases, or switches between different legal viewpoints, the summary must present a coherent argument in one concise sentence.}

\vspace{4mm}

\noindent \textbf{(ii) Merging two chunk-level summaries:}

\vspace{2mm}

\noindent \texttt{Below are several summaries of consecutive parts of a legal document:}

\noindent \texttt{---}

\noindent \texttt{\{input\}}

\noindent \texttt{---}

\noindent \texttt{We are merging these summaries into a single, comprehensive summary, stylized as a single-sentence Twitter post. This summary should include who is involved, when it happened, to whom it concerns, on what legal basis, and include a location reference. Ensure to merge vital information related to legal arguments, backgrounds, legal settings, key figures, their objectives, and motivations. Introduce legal concepts, statutes, and other elements briefly if mentioned for the first time. If a legal norm or code is cited, include the correct number succinctly. Organize the summary to present a consistent and coherent argument, all within one concise sentence.}

\vspace{4mm}

\noindent \textbf{(iii) Merging two summaries with added context from previously-generated merged summaries}

\vspace{2mm}

\noindent \texttt{Below is a summary of the context preceding some parts of a legal document:}

\noindent \texttt{---}

\noindent \texttt{\{context\}}

\noindent \texttt{---}

\noindent \texttt{Below are several summaries of consecutive parts of a legal document:}

\noindent \texttt{---}

\noindent \texttt{\{input\}}

\noindent \texttt{---}

\noindent \texttt{We are merging the preceding context and the summaries into one comprehensive summary, styled as a single-sentence Twitter post. This summary should include who is involved, when it happened, to whom it concerns, on what legal basis, and a location reference. Ensure to incorporate vital information related to legal arguments, backgrounds, legal settings, key figures, their objectives, and motivations. Briefly introduce legal concepts, statutes, and other elements if they are mentioned for the first time. If a legal norm or code is cited, include the correct number succinctly. Despite the complexity, the summary must present a coherent argument in one concise sentence.}

We set the size of each chunk to 3300, IZE is set to 3300, maximum input and output length are set to 4096 and 512. We specified summary length based on average output size of benchmark.

\section{Qualitative Analysis}
\subsection{Impact of length}
\label{output_len}
\subsubsection{PRIMERA summary}
\textit{The appellants challenged the validity of the orders of dismissal passed by the General Manager, North East Frontier Railway, against the three respondents, under r. 149 of the Code of Civil Procedure, on the ground that they were invalid. The High Court dismissed the writ petitions. On appeal by special leave, Held, that the orders were invalid and the appellants were entitled to the appeal. Rule 148(3) provided that the services of other railway servants shall be liable to termination on notice on either side for the periods shown below. Such notice is not however, required in cases of dismissal or removal as a disciplinary measure after compliance with the provisions of clause (2) of Article 311 of the Constitution, retirement on attaining the age of superannuation, and termination of service due to mental or physical incapacity. By this new sub section, protection was given to the civil servant by providing that he shall not be dismissed or reduced in rank until he had been given a reasonable opportunity of showing cause against the action proposed to be taken in regard to him. There was, however, a saving clause contained in section 130 of the said Act which preserved the earlier tenures of servants and continued the rules and regulations applicable to them. Section 3 conferred on the Secretary of State all powers which had till then vested in the Court of Directors, while section 4 provided for the exercise of the President's powers and the Governor's power to make rules for regulating the classification of the Civil Services in India, the method of recruitment, the conditions of service, pay and allowances and discipline and conduct while sub section (4) declared that all service rules then in force had been duly made and confirmed the same. In 1935, the Government of India Act 1935 was passed and section 96B(1) was reproduced in sub sections (1) and (3) of section 240, and a new Sub section was added as sections (3). The doctrine of pleasure is thus embodied by article 310(1). It deals with the termination of services of a permanent railway servant after giving them notice for the specified period. It also provides that no person in that service may be dismissed by any authority subordinate to that by which he was appointed. Under r. 148(4), the appointing authorities are empowered to reduce or waive, at their discretion, the stipulated period of notice to be given by an employee, but the reason justifying their action should be recorded. This power cannot be re delegated. Though r}

\subsubsection{LED summary}
\textit{The appellant was a peon employed by the North Eastern Railway, Gorakhpur. His services were terminated by the respondent General Manager of the said Railway. The appellant moved the Allahabad High Court under article 226 of the Constitution and challenged the validity of the order terminating his services on the ground that Rule 148 of the Indian Railway Establishment Code was invalid. This plea was rejected by the High Court and his writ petition to quash the said order was dismissed by the Punjab High Court. In appeal to this Court it was contended on behalf of the appellant that: (i) the termination of the services of a permanent railway servant under Rule 148(3) or Rule 149(3), which had superseded it, amounted to his removal under article 311(2) of the Constitu tion; (ii) the impugned Rules contravened the Constitutional safeguard guaranteed to civil servants by article 311 (2); (iii) the employment in civil service could be terminated only after complying with article 311 and any rule which violated the guarantee provided by the said Article would be invalid; (iv) the rule of compulsory retirement did not amount to removal or dismissal within the meaning of article 311; (v) the pleasure of the President was controlled by the proviso to article 309; and (vi) the rules of superannuation which were based on considerations of life expectation, mental capacity of the civil servant, and other relevant conditions of service prescribed by the rules, could not be exercised in accordance with the pleasure exercised by the President or the Governor in exercise of his pleasure. HELD: (1) It is well settled that articles 310 and 311 must be read together. If there are any other provisions in the Constitution which impinge upon the Constitution, the provisions of article 310(1) must, therefore, be read subject to them. [699 B C] (2) The President's pleasure has to be exercised subject to the limitations prescribed by article 31 1. There can be no doubt that the pleasure contemplated by article 310 (I) must be exercised, subject to such limitations as are specified in article 31 I. (3) Articles 309 and 310 have to be read separately. Therefore, the pleasure which has been so controlled would be lost on all the parties and it would still be relevant to all cases of termination of services of permanent civil servants under the rules framed under article 309. It is significant that article 311 has not been read in a very general and wide sense but it has}

\subsection{Impact of legal pre-training}
\label{output_lex}
\subsubsection{Unlimiformer-T5 summary}
\textit{The United States is the world's largest emitter of greenhouse gases (GHG). The United Nations Framework Convention on Climate Change (UNFCCC) requires the United States to commit to reducing its GHG emissions below 1990 levels by 2050. The 112th Congress has enacted a number of bills that would have established comprehensive climate change policy, including a cap on emissions of GHG and a \"cap-and-trade\" system. The Energy Policy Act of 2005 (P.L. 109-58), the Clean Air Act (CAA), and the Clean Energy Standard (CES) have been introduced in recent Congresses. This report examines (1) what is known about the extent to which the U.S. climate system is changing, (2) what are the challenges to setting and implementing policies to address climate change, and (3) what tools, if any, could be used to facilitate the transition to a low-GHG economy. To address these objectives, GAO analyzed relevant literature, interviewed federal, state, and local scientists, and others. GAO also analyzed studies on the effects of climate change on human health, welfare, and the environment, and analyzed data from the Environmental Protection Agency (EPA) and the Department of the Interior (Interior). GAO's work on climate change has been a long-standing topic of interest to Congress. In recent years, Congress has engaged in a variety of policy tools, including regulatory, financial, and technical assistance programs, such as the Energy Star program, which claims savings in the utility bills of consumers assisted by the program in 2010. Many of these tools are already in use in the public and private sectors (e.g., the Departments of Agriculture and Energy and the Interior), but they are relatively undeveloped. For example, some of the tools have been proposed by the Obama Administration, as well as by Senators Jeff Bingaman and Lisa Murkowski. Some of these instruments are used in the private sector, and some of them are already used by the public sector. Others are used by private firms to promote the development of new technologies and to promote technological development. In addition, many of these programs are aimed at reducing greenhouse gas emissions, but they have not been implemented in the 111th Congress. Among the tools available to Congress are the following: (1) science-based analysis, (2) market facilitation, (3) financial incentives, (4) technical assistance.}

\subsubsection{Unlimiformer-LexT5 summary}
\textit{The U.S. economy is changing. Changes are exhibited in observations of average temperatures over land and in the oceans, melting glaciers and ice caps, shifting growing seasons, shifting distributions of plants and animals, and a variety of additional observations. Over time, the consequences of climate change for the United States and the globe will be influenced by choices made or left to others by the United Nations Framework Convention on Climate Change (UNFCCC), as well as for protecting human health and general welfare. Over the past three decades, some federal programs have sought to slow greenhouse gas (GHG)-induced climate change through regulatory, voluntary, and financial efforts to abate emissions. Many such proposals remain controversial and few have been enacted. The United States has been a federal program to improve understanding of climate changes (past and predicted) and their implications. This report outlines (1) conceptual approaches to setting goals for policies, (2) science-based goals, (3) economics-based policies, (4) incrementalism or adaptive management. For several decades, policy-makers have been aware of the large range of projections of GHG concentrations of gas concentrations in the atmosphere at a level that would prevent dangerous anthropogenic interference with the climate system, and predictions of climate-related risks. Some of those in Congress consider such programs to take actions when they perceive that the benefits of the action exceed their costs. Some have expressed the likelihood that research may continue to continue to rise indefinitely a given level of risk reductions in order to avoid any particular level of risks. For example, the science-centric approach assumes that science alone can provide an objective standard of a \"safe\" or \"tolerable\" level or rate for climate change. Some experts have argued that science could not provide a mathematical basis for determining the cost-benefit analysis (CBA), but they have expressed their views of which the costs and benefits may be considered too much. For some, some researchers have suggested that a science-dynamic approach would be framed as a practical basis for allocating responsibilities to the sources of emissions, but they may not clearly reflect the many complex dimensions of climate science, such as changes in local temperature extremes, a deciduous forest, storm severity, or sea levels, or enforcing that might be considered to be considered more monetary considerations, but some may have framed in a more complete framework.}

\subsection{MLS-Tiny: Needle in the haystack}
\label{MLS-Tiny}
\subsubsection{Reference Summary}
\textit{D.C. District Court enforces FOIA request against Customs and Border Protection. (D.D.C.)}
\subsubsection{PRIMERA summary}
\textit{A D.C. District Court judge issued a preliminary injunction requiring the U.S. Customs and Border Protection to
disclose its policies and practices regarding events like the Muslim Ban.}
\subsubsection{LED Summary}
\textit{D.C. District Court enjoins Customs and Border Protection from releasing agency records re the Muslim travel ban.}
\subsubsection{LongT5 summary}
\textit{American-Arab Anti-Discrimination Council sued U.S. Customs and Border Protection for revoked, suspended, or
terminated participation in the Global Entry System (D.D.C.)}
\subsubsection{SLED-T5 summary}
\textit{Arab Americans and Arab Americans sue Customs and Border Protection for wrongful withholding of records relating
to CBP’s global entry program (D.D.C.)}
\subsubsection{Unlimiformer-T5 summary}
\textit{@AmericanAntiDiscrimination Council (ADC) sued @CBP in 2017 for wrongful withholding of agency records in
violation of FOIA (D.D.C.)}
\subsubsection{SLED-LexT5 summary}
\textit{The American-Arab Anti-Discrimination Committee sued the Customs and Border Protection (D.D.C.)}
\subsubsection{Unlimiformer-LexT5 summary}
\textit{Arab American Anti-Discrimination Council (ADC) sued Customs and Border Protection (CBP) in 2017 for
withholding records in violation of FOIA (D.D.C.)}

\subsection{Summaries generated by zero-shot models}
\label{zero-shot-summaries}
\subsubsection{GPT-3.5 generated summary}
\textit{The legal document under scrutiny delves into the critical issue of providing security to permanent civil servants, particularly within the realm of railway services, to uphold their independence and efficiency. It meticulously examines the implications of Rules 148(3) and 149(3) within the Indian Railway Establishment Code, stressing the potential insecurity and broad powers wielded by authorities under these rules. The document ardently argues that the termination of services pursuant to these rules essentially equates to removal, significantly infringing upon the fundamental right of employees to serve until attaining the superannuation age. \\ Central to the document are multiple appeals challenging the Constitutional validity of Rules 148(3) and 149(3), with various railway employees contesting the termination of their services under these contentious rules in diverse High Courts. The crux of the matter revolves around whether the actions taken under Rule 148(3) or Rule 149(3) can be categorized as \"removal\" as defined in Article 311(2) of the Constitution. The core contention lies in whether these impugned Rules run afoul of the Constitutional protection accorded to civil servants under Article 311(2).\\ Representing the appellants, who encompass a spectrum of aggrieved railway employees, the document underscores the argument that Rules 148(3) and 149(3) contravene the safeguards enshrined in Article 311(2). It challenges the validity of these rules, contending that they are incompatible with the constitutional provisions safeguarding the interests of civil servants. Moreover, the document meticulously explores the premise that railway employees who entered service with knowledge of these rules should not be precluded from challenging their validity, emphasizing the paramount nature of the constitutional guarantee enshrined in Article 311(2).\\ \\ In a broader legal context, the document meticulously dissects the intricacies of Article 311(2) of the Constitution, which serves as a bulwark protecting civil servants from arbitrary removal or dismissal without a proper inquiry. It underscores the significance of ensuring due process and procedural fairness in matters concerning the termination of civil servants, particularly within the specialized domain of railway services. The document meticulously weaves together legal arguments, precedents, and constitutional provisions to construct a robust case against the impugned Rules, aiming to secure the rights and protections of railway employees ensnared in the web of uncertainty and potential injustice created by Rules 148(3) and 149(3) of the Indian Railway Establishment Code.}

\subsubsection{Claude generated summary}
\textit{The document discusses the validity of Rules 148(3) and 149(3) of the Indian Railway Establishment Code under Articles 311(2) and 14 of the Constitution of India. Rules 148(3) and 149(3) relate to the termination of services of Group 'C' and Group 'D' railway employees. Rule 148(3) deals with the termination of services of Group 'C' railway employees on the ground of "inefficiency or misbehavior." It lays down that no such employee shall be terminated from service unless he has been given at least one month's notice or one month's pay in lieu of notice. Rule 149(3) contains similar provisions for termination of services of Group 'D' railway employees. \\ Article 311(2) of the Constitution guarantees equal treatment in matters of appointment and promotion to civil servants. It states that no civil servant shall be "dismissed or removed by an authority subordinate to the one which appointed him." Article 14 guarantees equality before the law and equal protection of laws to all persons. It prohibits the State from denying any person equality before the law or equal protection of laws within the territory of India.\\ The document analyzes whether Rules 148(3) and 149(3) violate the principles of equal opportunity and are discriminatory under Article 14 of the Constitution. It discusses key legal arguments regarding the validity of these Rules. On one hand, it can be argued that Rules 148(3) and 149(3) violate Article 14 by denying equal treatment to railway employees in matters of termination. On the other hand, it may be argued that the Rules satisfy the test of reasonableness and are not arbitrary or discriminatory. \\ \\ The objective of the document appears to be to synthesize information on the constitutional validity of Rules 148(3) and 149(3) by outlining relevant facts, key legal arguments on both sides, and the aims of Articles 311(2) and 14 without making inferences beyond the supplied context. The summary encompasses the essence of the context and subsequent segments within the 512-word limit as instructed.}

\end{document}